\documentclass[10pt,journal,compsoc]{IEEEtran}
\ifCLASSOPTIONcompsoc
  \usepackage[nocompress]{cite}
\else
  \usepackage{cite}
\fi

\ifCLASSINFOpdf
\else
\fi

\usepackage[colorlinks, citecolor=green]{hyperref}
\usepackage{graphicx}
\usepackage{amssymb}
\usepackage{amsmath}
\usepackage{amsfonts}
\usepackage{subfig}
\usepackage{array}
\usepackage{booktabs}
\begin{document}
\title{Image Projective Invariants}

\author{Erbo~Li$^{*}$,
        Hanlin~Mo$^{*}$,
        Dong~Xu,
        Hua~Li,~\IEEEmembership{Senior Member,~IEEE}% <-this % stops a space
\IEEEcompsocitemizethanks{
\IEEEcompsocthanksitem $^{*}$ Denotes equal contribution.
\IEEEcompsocthanksitem E. Li is with EON Reality Inc, Irvine, CA, 92618.\protect\\
Email: sophialiuli@gmail.com
\IEEEcompsocthanksitem H. Mo (correspondent author) and H. Li are with Key lab of Intelligent Information Processing, Institute of Computing Technology, Chinese Academy of Sciences, 100190 Beijing; University of Chinese Academy of Sciences, 100049 Beijing. E-mail:\{mohanlin, lihua\}@ict.ac.cn
\IEEEcompsocthanksitem D.Xu is with Ambry Genetics, Aliso Viejo, CA 92656.\protect\\
Email: xudong0614@hotmail.com}}

\IEEEtitleabstractindextext{%
\begin{abstract}
In this paper, we propose relative projective differential invariants ($RPDIs$) which are invariant to general projective transformations. By using $RPDIs$ and the structural frame of integral invariant, projective weighted moment invariants ($PIs$) can be constructed very easily. It is first proved that a kind of projective invariants exists in terms of weighted integration of images, with relative differential invariants as the weight functions. Then, some simple instances of $PIs$ are given. In order to ensure the stability and discriminability of $PIs$, we discuss how to calculate partial derivatives of discrete images more accurately. Since the number of pixels in discrete images before and after the geometric transformation may be different, we design the method to normalize the number of pixels. These ways enhance the performance of $PIs$. Finally, we carry out some experiments based on synthetic and real image datasets. We choose commonly used moment invariants for comparison. The results indicate that $PIs$ have better performance than other moment invariants in image retrieval and classification. With $PIs$, one can compare the similarity between images under the projective transformation without knowing the parameters of the transformation, which provides a good tool to shape analysis in image processing, computer vision and pattern recognition.
\end{abstract}

\begin{IEEEkeywords}
General projective transformations, relative projective differential invariants, projective weighted moment invariants, object recognition.
\end{IEEEkeywords}}

\maketitle

\IEEEdisplaynontitleabstractindextext

% \fi
\IEEEpeerreviewmaketitle

\begin{figure}
  \centering
  \includegraphics[height=40mm,width=80mm]{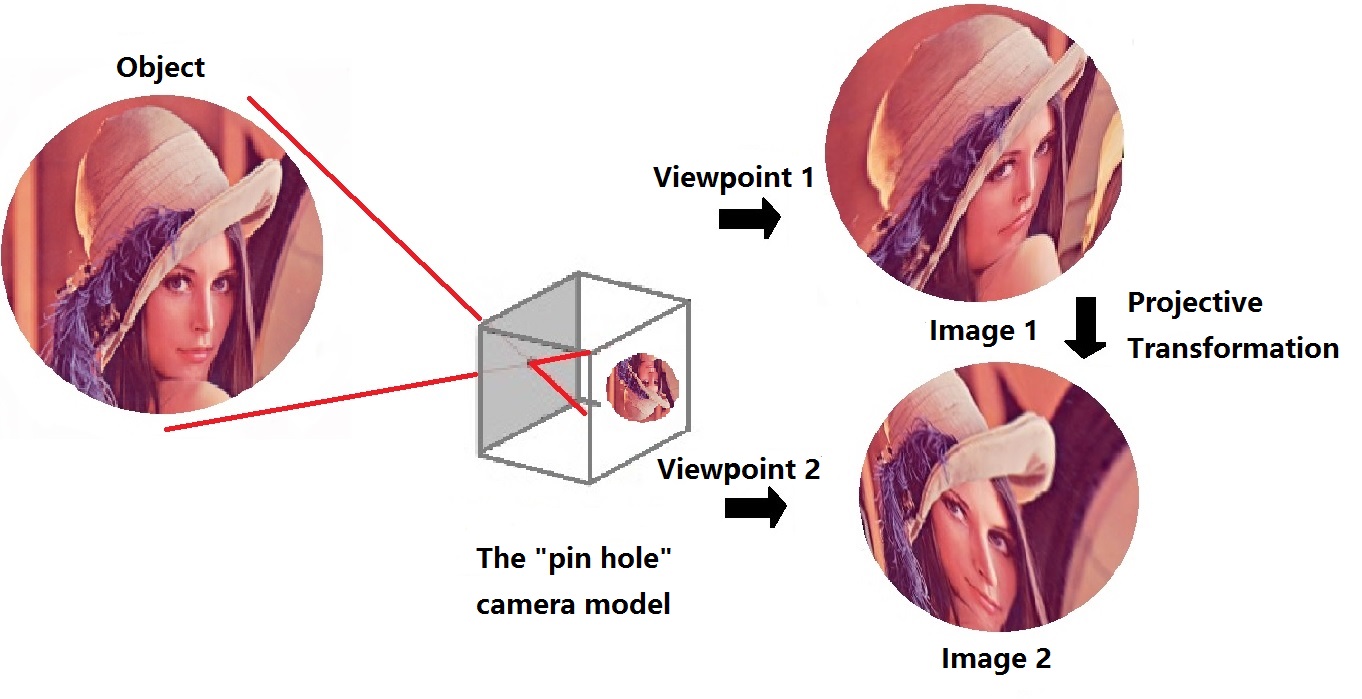}\\
  \caption{Projective transformation from a pin-hole camera model}\label{Fig:1}
\end{figure}

\IEEEraisesectionheading{\section{Introduction}\label{sec:1}}
\IEEEPARstart{H}{ow} to measure the similarity or difference of a scene or an object observed from different viewpoints is one of fundamental problems in computer vision and object recognition. As shown in Fig.~\ref{Fig:1}, a general camera model is the "pin-hole", a projection of projective transformation. The severe geometric deformation caused by projective transformation brings some difficult to judge whether the two images of a 3D scene contain the same object. So, it's necessary to construct a kind of image features which are invariant to general projective transformations.

Geometric moment invariant is a good idea, which was first introduced into computer vision community by \cite{5}, using moments to define invariants of images and shapes. The definition of moment has a deep background in physics and mathematics. Some geometric moment invariants under transformations of translation, rotation and scaling, or even affine were built up and have been widely used in applications as global features \cite{2,9,18,30}. For the case of small objects with respect to large distance of camera-to-scene, the effect of projective deformation is slight, which can be approached by affine transformation as usual. For computer vision tasks like robotic vision and object recognition that require precise calculation, the effect of projective deformation can no longer be neglected. In the light of the classical theory of geometric moment invariants, it seems easy to seek for projective invariants in the similar way. But in fact, a major problem occurs that the projective transformation is not linear in transform parameters, whose Jacobian performs as a function of the coordinates instead of a constant. Therefore, such quality determines the fact that the traditional way of generating geometric invariants is not longer valid. In order to adapt the general solution to projective transformation, a possible way goes to reconsider the structure of traditional geometric invariants to combine the structural method with the essence of projective transformation. For general projective transformations, \cite{17} attempted to express projective invariants in the form of infinite series of moments. Unfortunately, the definition setting was problematic and unwarranted. The experimental results presented in \cite{17} were also unsatisfactory. \cite{19, 22, 23, 24} gave some sorts of projective invariants under strict constraints or subsets, which lose the generality and therefore can't to extend for broader applications. It is worth noting that the determinant consisting of partial differentials was designed in \cite{25} which can be used to construct general projective moment invariants. Also, they defined differential moments which are named as D-moments. But we will point out that one of the structural formulas proposed in \cite{25} was wrong. Therefore, all experiments based on this formula didn't have reference value.

In addition, the studies of differential invariants should be concerned. Unlike moment invariants constructed by integral frames, differential invariants are not global features but local features. Theoretically, they have good invariance to interference such as occlusion. So far, differential invariants have been studied mainly for affine transformations. In \cite{13}, Olver constructed affine differential invariants by using the moving frame method. Then, he found that two kinds of affine differential invariants can be use to define the affine gradient \cite{14}. Wang et~al. presented a new method to derive a special type of affine differential invariants \cite{21}. Given some functions defined on the plane and the affine group acting on the plane, there were induced actions of the group on the functions and their derivative functions. But we must point out that these methods are not easy to use. Recently, Li et~al. found the isomorphism between differential invariants and geometric moment invariants to general affine transformations \cite{7}. If affine moment invariants were known, relative affine differential invariants can be obtained by the substitution of moments by partial derivatives with the same order. This method made the construction of affine differential invariants very easy. Also, they gave a construction formula of relative projective differential invariants. In \cite{10}, Mo et~al. combined affine differential invariants with affine moment invariants, and obtained affine weighted moment invariants. Compared with the traditional moment invariants, this kind of invariants achieved better results in image retrieval due to the use of both local and global information. This method provides a new way for our research.

The main contributions of this paper are summarized as follows.
\begin{enumerate}
  \item The projective differential invariants are explicitly reported, which are invariant under general projective transformations.
  \item It is proved that a kind of projective weighted moment invariants ($PIs$) exist in terms of weighted integration of images, with $RPDIs$ as the weight functions.
  \item The reasons which cause $PIs$ to be calculated imprecise are analyzed. The appropriate method to calculate partial derivatives of discrete images is selected. Also, the normalized method is used to deal with the change in the number of pixels. These ways make $PIs$ have practical value.
  \item Experimental results obtained by using the projective invariants based on correct structural formulas are obtained for the first time. By comparing with other moment invariants, it's obviously that $PIs$ have advantages.
\end{enumerate}

In Section~\ref{sec:2}, some definitions and notations are given. Then, we introduce some related works and point out their mistakes and limitations in Section~\ref{sec:3}. Section~\ref{sec:4},~\ref{sec:5} are the major parts of this paper. We give the definitions of $RPDIs$ and the structural framework of $PIs$. Experiments and discussions are shown in Section~\ref{sec:6}. At last, some conclusions are given in Section~\ref{sec:7}.

\section{Basic Definitions and notations}\label{sec:2}
In this section, we will introduce some basic definitions and notations that will be used to construct $RPDIs$ and $PIs$.
\subsection{Invariants and Integral Invariants}
Given a set of parameters $(a, b, c, ...)$, the transformed parameters are $(\alpha, \beta, \gamma, ...)$ under the transformation T with the correspondence of $a$ to $\alpha$, $b$ to $\beta$, etc. If there is a function $I$ satisfying (1), then $I$ is called relative invariant.
\begin{equation}\label{equ:1}
I(\alpha, \beta, \gamma, ...)=W^{k} \cdot I(a,b,c,...)
\end{equation}
when the power $k=0$, $I$ is called absolute invariant. Integral invariants are defined in the form of multiple integrations \cite{5,30}. It's crucial to find correct integral cores remaining the same constructions under corresponding transformations. Then, they are used to calculate the multiple integrations in (\ref{equ:2}), where $c_{i}$ is the integral core, $i=1,2,..$.
\begin{equation}\label{equ:2}
  I=\int...\int c_{1}c_{2}...c_{n}dx_{1}dy_{1}dx_{2}dy_{2}...dx_{n}dy_{n}
\end{equation}

\subsection{General Projective Transformation}\label{sec:2.2}
A general projective transformation between $2D$ points $(x,y)$ and $(u,v)$ is defined by (\ref{equ:3}).
\begin{equation}\label{equ:3}
u=\frac{ax+by+c}{px+qy+r}~~~~~v=\frac{dx+ey+f}{px+qy+r}
\end{equation}
where all parameters are real, and $p^{2}+q^{2} \neq 0$. If both $p$ and $q$ are zero, it reduces to an affine transformation. Notice that there are totally 9 parameters in (\ref{equ:3}). Since the numerators and the denominators can be divided by a nonzero constant, a common way to simplify (\ref{equ:3}) is to let r equal to constant 1 to eliminate the parameter without losing the generality. Therefore, there are 8 independent parameters for a general projective transformation which includes some other transformations as its special cases. Let A represent the coefficient determinant of (\ref{equ:3}).
\begin{equation}\label{equ:4}
A=
\begin{vmatrix}
a&b&c\\
d&e&f\\
p&q&r
\end{vmatrix}
\end{equation}

The Jacobian determinant J of the projective transformation (\ref{equ:3}) is
\begin{equation}\label{equ:5}
 J=J(x,y)=
\begin{vmatrix}
u_{x}&u_{y}\\
v_{x}&v_{y}
\end{vmatrix}
=\frac{A}{(px+qy+r)^{3}}
\end{equation}

\subsection{Moment and Weighted Moment}
For an image function $f(x,y)$, the $(i+j)$ order geometric moment is defined as an double integral
\begin{equation}\label{equ:6}
  m_{ij}=\iint x^{i}y^{j}f(x,y)dxdy
\end{equation}

If there is another weight function besides the image function $f(x,y)$ in the integrand, it is called weighted moment which is defined by
\begin{equation}\label{equ:7}
 M_{ij}=\iint x^{i}y^{j}f(x,y)w(x,y)dxdy
\end{equation}
where $w(x,y)$ is called the weight function.

\section{Related work}\label{sec:3}
From the viewpoint of projective geometry, the only invariant property for general projective transformations is the cross ratio, which can be expressed in several ways. The cross ratio is defined locally for points on a straight line or line bundles and is not easy to directly be applied to images \cite{11}. As a result, the researchers began to construct new projective invariants of images and achieved some results. In this section, some previous work directly related to this paper will be described. Also, we will point out their limitations and mistakes.
\subsection{Restricted Projective Transformation}
Because of the complex structure of general projective transformations, some researchers focused on restricted projective transformations firstly. In \cite{19}, Voss and Susse defined a kind of finite projective invariant by
\begin{equation}\label{equ:8}
R_{pq}=\iint \frac{x^{p}y^{q}}{p^{\alpha}(x,y)}f(x,y)dxdy
\end{equation}
where $p(x,y)=\sum_{i+j=n}p_{ij}x^{i}y^{j}$, $p_{ij}$ is coefficient, $\alpha=\frac{(p+q+3)}{n}$ and $n>0$.

These invariants fit for a special case of general projective transformations called "rein transform" with $b=c=d=f=0$, which means
\begin{equation}\label{equ:9}
  u=\frac{ax}{px+qy+r}~~~~~v=\frac{ey}{px+qy+r}
\end{equation}

In fact, it is obviously that (\ref{equ:8}) is a kind of weighted moments, with $\frac{1}{p^{\alpha}(x,y)}$ as the weight function.

\cite{22, 23, 24} gave some new results. In \cite{22}, Wang et.al extended the moment definition to allow the power of coordinates varies from non-negative integers to arbitrary integers and gave moment-like invariants in rational form for a special case defined by (\ref{equ:10}). This kind of invariants was constructed by (\ref{equ:11}), where $n$ is a positive even integer and $k$ is a positive integer.
\begin{equation}\label{equ:10}
  u=\frac{ax}{px+qy+r}~~~~~v=\frac{dx+ey}{px+qy+r}
\end{equation}
\begin{equation}\label{equ:11}
 I(n,k)=\frac{m^{2(k-1)}_{-3,0}\sum^{kn}_{i=0}(-1)^{i}\tbinom{kn}{i}m_{-i-3,i}m_{i-kn-3,kn-i}}{(\sum_{i=0}^{n}(-1)^{i}\tbinom{n}{i}m_{-i-3,i}m_{i-n-3,n-i})^{k}}
\end{equation}

\cite{24} proposed co-moment to construct projective invariants, under the condition that the correspondence of two reference points in images was known beforehand. It was said to be the first paper to establish a set of easily implemented projective invariants for 2D images. But this method is not free, independent technique and relying on the two correspondent points seriously restricts its application. For example, it is not easy for image retrieval in large image database, the amount of calculation is unconceivable for comparison between the given image and database images.

\subsection{General Projective Transformation}\label{sec:3.2}
Comparatively speaking, the number of studies conducted on general projective transformations is much less. By using Lie Group theory, \cite{20} proved that there are no finite projective invariants. \cite{17} gave another proof by decomposing the general projective transformation (\ref{equ:3}) into eight one-parameter transformations. In fact, the real meaning of those proofs can be understood as that there are no simple or direct projective moment invariants. Other forms of invariants or weighted moment invariants may still be possible. Suk and Flusser tried to extend their work on affine moment invariants to projective moment invariants in \cite{17}. They noticed that the determinant of three points $(x_{i},y_{i}), i=1, 2, 3$
\begin{equation}\label{equ:12}
d(1, 2, 3)=
\begin{vmatrix}
x_{1}&x_{2}&x_{3}\\
y_{1}&y_{2}&y_{3}\\
1&1&1
\end{vmatrix}
\end{equation}
would be changed to $d^{'}(1, 2, 3)$, if $(x_{i},y_{i})$ was transformed into $(u_{i},v_{i})$, $i=1,2,3$ by the projective transformation (\ref{equ:3}).
\begin{equation}\label{equ:13}
d^{'}(1, 2, 3)=
\begin{vmatrix}
u_{1}&u_{2}&u_{3}\\
v_{1}&v_{2}&v_{3}\\
1&1&1
\end{vmatrix}
\end{equation}

\begin{equation}\label{equ:14}
\begin{split}
d^{'}(1,2,3)&=\frac{A\cdot d(1,2,3)}{(px_{1}+qy_{1}+1)(px_{2}+qy_{2}+1)(px_{3}+qy_{3}+1)}\\
&=\sqrt[3]{J(x_{1},y_{1})}\sqrt[3]{J(x_{2},y_{2})}\sqrt[3]{J(x_{3},y_{3})}d(1,2,3)
\end{split}
\end{equation}

The relationship (\ref{equ:14}) was used to construct a kind of infinite projective moment invariants, which can be expressed as the form of infinite series of moments. The basic idea was that the Jacobian determinant in the transformation (\ref{equ:3}) contains a denominator of power 3 as in (\ref{equ:5}). If the frequency of each points appearing in the definition of moment integration is exactly three, the Jacobian determinants and the denominator in (\ref{equ:14}) would be canceled out. In this way, given three points $(x_{1},y_{1}), (x_{2}, y_{2})$ and $(x_{3},y_{3})$ in the original image, the projective moment invariants can be defined by
\begin{equation}\label{equ:15}
 \int...\int\frac{f(x_{1},y_{1})f(x_{2},y_{2})f(x_{3},y_{3})}{d^{3}(1,2,3)}dx_{1}dy_{1}dx_{2}dy_{2}dx_{3}dy_{3}
\end{equation}
where $f(x_{i},y_{i})$ is the image intensity, $i=1, 2, 3$. It is easy to prove that (\ref{equ:15}) is a kind of projective invariants. By expanding $d(1, 2, 3)$ as power series of $x_{i}y_{i}$, Suk and Flusser got an infinite moment series which was called infinite projective invariant. More points involvements were allowed with the definition. They also got two instance of (\ref{equ:15}) by setting the point number $N=3, 4, 5$.

However, two problems should be noted here.
\begin{enumerate}
\item The infinite projective invariant was defined in moment, its form was popular and the calculation was straightforward. But this method was difficult to use. Its error limit was hard to evaluate and the calculation may be time-consuming. We had to compute a large number of moments to ensure that the invariants are stable.
\item More importantly, Suk and Flusser gave three instances by setting the point number $N=3, 4, 5$, but only $N=4$ is correct. For $N=3, 5$, the invariants are always zero.
\end{enumerate}

The second problem can be explained simply as following. When we exchange the order of integration, the final result of (\ref{equ:15}) does not change. So, when $N=3$, we have
\begin{equation}\label{equ:16}
\begin{split}
I&=\int...\int\frac{f(x_{1},y_{1})f(x_{2},y_{2})f(x_{3},y_{3})}{d^{3}(1,2,3)}dx_{1}dy_{1}dx_{2}dy_{2}dx_{3}dy_{3}\\
&=\int...\int\frac{f(x_{1},y_{1})f(x_{2},y_{2})f(x_{3},y_{3})}{d^{3}(1,2,3)}dx_{2}dy_{2}dx_{1}dy_{1}dx_{3}dy_{3}
\end{split}
\end{equation}

Meanwhile, after the change of variables, we can obtain
\begin{equation}\label{equ:17}
\begin{split}
I&=\int...\int \frac{f(x_{1},y_{1})f(x_{2},y_{2})f(x_{3},y_{3})}{d^{3}(1,2,3)}dx_{1}dy_{1}dx_{2}dy_{2}dx_{3}dy_{3}\\
&=\int...\int \frac{f(x_{2},y_{2})f(x_{1},y_{1})f(x_{3},y_{3})}{d^{3}(2,1,3)}dx_{2}dy_{2}dx_{1}dy_{1}dx_{3}dy_{3}
\end{split}
\end{equation}

And from the property of determinant, the change of point order will change its sign.
\begin{equation}\label{equ:18}
  d^{3}(2,1,3)=(-1)^{3} \cdot d^{3}(1,2,3)
\end{equation}

Thus, by using (\ref{equ:16}), (\ref{equ:17}) and (\ref{equ:18}), we can find that
\begin{equation}\label{equ:19}
  I=(-1)^{3}I=-I
\end{equation}

When $N=5$, the reason is similar. Therefore, the experimental result didn't have reference value, which was obtained by using the invariant $(N=3)$ in \cite{18}.

Recently, Wang et~al. proposed a kind of projective invariants in \cite{25}. Let an image $f(x,y)$ be transformed by (\ref{equ:3}) into the image $g(u,v)$. $(u_{1},v_{1}), (u_{2},v_{2})$ and $(u_{3},v_{3})$ in $g(u,v)$ are the corresponding points of $(x_{1},y_{1}),(x_{2},y_{2})$ and $(x_{3},y_{3})$ in $f(x,y)$. Suppose that both $f(x,y)$ and $g(u,v)$ have the first-order partial derivatives.

Then, they defined two determinants by
\begin{equation}\label{equ:20}
\begin{split}
&D(1,2,3)=\\
&\begin{vmatrix}
x_{1}\frac{\partial{f}}{\partial{x_{1}}}+y_{1}\frac{\partial{f}}{\partial{y_{1}}}&x_{2}\frac{\partial{f}}{\partial{x_{2}}}+y_{2}\frac{\partial{f}}{\partial{y_{2}}}&x_{3}\frac{\partial{f}}{\partial{x_{3}}}+y_{3}\frac{\partial{f}}{\partial{y_{3}}}\\
\frac{\partial{f}}{\partial{x_{1}}}&\frac{\partial{f}}{\partial{x_{2}}}&\frac{\partial{f}}{\partial{x_{3}}}\\
\frac{\partial{f}}{\partial{y_{1}}}&\frac{\partial{f}}{\partial{y_{2}}}&\frac{\partial{f}}{\partial{y_{3}}}
\end{vmatrix}
\end{split}
\end{equation}

\begin{equation}\label{equ:21}
\begin{split}
&D^{'}(1,2,3)=\\
&\begin{vmatrix}
u_{1}\frac{\partial{g}}{\partial{u_{1}}}+v_{1}\frac{\partial{g}}{\partial{v_{1}}}&u_{2}\frac{\partial{g}}{\partial{u_{2}}}+v_{2}\frac{\partial{g}}{\partial{v_{2}}}&u_{3}\frac{\partial{g}}{\partial{u_{3}}}+v_{3}\frac{\partial{g}}{\partial{v_{3}}}\\
\frac{\partial{g}}{\partial{u_{1}}}&\frac{\partial{g}}{\partial{u_{2}}}&\frac{\partial{g}}{\partial{u_{3}}}\\
\frac{\partial{g}}{\partial{v_{1}}}&\frac{\partial{g}}{\partial{v_{2}}}&\frac{\partial{g}}{\partial{v_{3}}}
\end{vmatrix}
\end{split}
\end{equation}

There is a relation
\begin{equation}\label{equ:22}
 D^{'}(1,2,3)= D(1,2,3)\cdot K
\end{equation}
where
\begin{equation}\label{equ:23}
\begin{split}
 K&=\frac{(px_{1}+qy_{1}+1)(px_{2}+qy_{2}+1)(px_{3}+qy_{3}+1)}{A}\\
 &=\frac{1}{\sqrt[3]{J(x_{1},y_{1})}}\frac{1}{\sqrt[3]{J(x_{2},y_{2})}}\frac{1}{\sqrt[3]{J(x_{3},y_{3})}}\\
\end{split}
\end{equation}

By using (\ref{equ:12}) and (\ref{equ:20}),  Wang et~al. constructed two kinds of projective invariants, which were defined by
\begin{equation}\label{equ:24}
\begin{split}
Inv_{1,n}=&\int...\int (d(1,2,3))^{n}(D(1,2,3))^{n+3}\\&dx_{1}dy_{1}dx_{2}dy_{2}dx_{3}dy_{3}
\end{split}
\end{equation}
\begin{equation}\label{equ:25}
\begin{split}
Inv_{2,n}=&\int...\int (d(1,2,3)d(1,2,4)d(1,3,4)d(2,3,4))^{n}\\&(D(1,2,3)D(1,3,4)D(1,2,4)D(2,3,4))^{n+1}\\&dx_{1}dy_{1}dx_{2}dy_{2}dx_{3}dy_{3}dx_{4}dy_{4}
\end{split}
\end{equation}
where $n=0,1,2,...$.

These projective invariants can be represented as polynomials of D-moment which was defined by
\begin{equation}\label{equ:26}
  \phi_{pqrst}=\iint x^{p}y^{q}(\frac{\partial{f}}{\partial{x}})^{r}(\frac{\partial{f}}{\partial{y}})^{s}(x\frac{\partial{f}}{\partial{x}}+y\frac{\partial{f}}{\partial{y}})^{t}f(x,y)dxdy
\end{equation}
where $p,q,r,s,t=0,1,2,...$. Obviously, (\ref{equ:26}) is a kind of weighted moments. Also, there are two problems we have to pay attention to.
\begin{enumerate}
  \item In \cite{25}, all experimental results were obtained by using the instances of $Inv_{1,n}$, when $n=0,1$. Unfortunately, similarly to (\ref{equ:15}), (\ref{equ:24}) is always zero. Because we can find that $I_{1,n}=(-1)^{2n+3}I(1,n)=-I_{1,n}$. Therefore, the results in \cite{25} are not valid.
  \item The definition of $Inv_{2,n}$ is theoretically correct. But, only $Inv_{2,0}$ can be used in practice. When $n=1$, the expansion of $Inv_{2,1}$ contains more than 70 million terms. And as $n$ grows, the number of terms increases exponentially. This means that (\ref{equ:25}) only constructs one projective invariant.
\end{enumerate}

In summary, the projective invariants that have been obtained with practical value are all for restricted projective transformations. So, their application scenarios are greatly limited. Two kinds of invariants for general projective transformations have obvious defects in theories. Thus, the problem of real projective invariants for general projective transformations has been being widely open.

\section{Relative Projective differential invariants}\label{sec:4}
In this section, we will give two definitions of $RPDIs$. One is proposed by Li et~al. in \cite{7}. Another is defined for the first time in this paper.

\hspace{-5mm}\textbf{Definition 1.}
Support that an image function $f(x,y)$ has the second-order partial derivatives. Then $RPDI_{1}$ can be defined by
\begin{equation}\label{equ:27}
RPDI_{1}(x,y)=\frac{\partial^{2}{f}}{\partial{x^{2}}}(\frac{\partial{f}}{\partial{y}})^{2}-2\frac{\partial{f}}{\partial{x}}\frac{\partial{f}}{\partial{y}}\frac{\partial^{2}{f}}{\partial{x}\partial{y}}+\frac{\partial^{2}{f}}{\partial{y^{2}}}(\frac{\partial{f}}{\partial{x}})^{2}
\end{equation}

\hspace{-5mm}\textbf{Theorem 1.}
Let an image $f(x,y)$ be transformed by (\ref{equ:3}) into the image $g(u,v)$. Suppose that $f(x,y)$ and $g(u,v)$ both have the second-order partial derivatives, then we have
\begin{equation}\label{equ:28}
  RPDI_{1}^{'}(u,v)= \frac{RPDI_{1}(x,y)}{J^{2}}
\end{equation}
where
\begin{equation}\label{equ:29}
  RPDI_{1}^{'}(u,v)=\frac{\partial^{2}{g}}{\partial{u^{2}}}(\frac{\partial{g}}{\partial{v}})^{2}-2\frac{\partial{g}}{\partial{u}}\frac{\partial{g}}{\partial{v}}\frac{\partial^{2}{g}}{\partial{u}\partial{v}}+\frac{\partial^{2}{g}}{\partial{v^{2}}}(\frac{\partial{g}}{\partial{u}})^{2}
\end{equation}

The proof of Theorem 1 is obvious by using $Maple$. It should be noted that $RPDI_{1}(x,y)$ has a geometric meaning. If an image $f(x,y)$ is taken as a "curved surface" $z=f(x,y)$ defined on 2D region, the traditional differential geometry methods can be applied on it. There are two movement invariants on curved surfaces in Euclidean space, Gaussian curvature $K(x,y)$ and mean curvature $H(x,y)$ \cite{3}. They are defined by
\begin{equation}\label{equ:30}
K(x,y)=\frac{\frac{\partial^{2}{f}}{\partial{x^{2}}}\frac{\partial^{2}{f}}{\partial{y^{2}}}-(\frac{\partial^{2}{f}}{\partial{x}\partial{y}})^{2}}{(1+(\frac{\partial{f}}{\partial{x}})^{2}+(\frac{\partial{f}}{\partial{y}})^{2})^{2}}
\end{equation}
\begin{equation}\label{equ:31}
H(x,y)=\frac{(1+(\frac{\partial{f}}{\partial{y}})^{2})\frac{\partial^{2}{f}}{\partial{x^{2}}}-2\frac{\partial{f}}{\partial{x}}\frac{\partial{f}}{\partial{y}}\frac{\partial^{2}{f}}{\partial{x}\partial{y}}+(1+(\frac{\partial{f}}{\partial{x}})^{2})\frac{\partial^{2}{f}}{\partial{y^{2}}}}{2(1+(\frac{\partial{f}}{\partial{x}})^{2}+(\frac{\partial{f}}{\partial{y}})^{2})^{\frac{3}{2}}}
\end{equation}

The numerator $N(x,y)$ of (\ref{equ:30}) is a Hessian determinant, and the numerator of $H(x,y)$ can be separated into two parts, $L(x,y)$ and $RPDI_{1}(x,y)$. $L$ is defined by
\begin{equation}\label{equ:32}
L(x,y)=\frac{\partial^{2}{f}}{\partial{x^{2}}}+\frac{\partial^{2}{f}}{\partial{y^{2}}}
\end{equation}

It's well known that Laplace descriptor $L(x,y)$ is a rotation invariant. In \cite{13}, Olver pointed out that the numerator $N(x,y)$ of (\ref{equ:30}) and $RPDI_{1}(x,y)$ in (\ref{equ:31}) were two relative affine differential invariants. With further analysis, Li et~al. found the interesting result that $RPDI_{1}(x,y)$ is also a relative projective differential invariant \cite{7}. Then, we will define a new structural formula of $RPDIs$.

\hspace{-5mm}\textbf{Definition 2.}
Support that an image $f(x,y)$ has the third-order partial derivatives. Then we have
\begin{equation}\label{equ:33}
\begin{split}
&RPDI_{2}(x,y)=\\&S^{2}(x,y)-12\cdot RPDI_{1}^{2}(x,y)\cdot N(x,y)-12 \cdot RPDI_{1}(x,y)\\& \cdot \{-(\frac{\partial{f}}{\partial{x}})^3\frac{\partial^{3}{f}}{\partial{x}\partial{y^{2}}}\frac{\partial^{2}{f}}{\partial{y^{2}}}+(\frac{\partial{f}}{\partial{x}})^{3}\frac{\partial^{2}{f}}{\partial{x}\partial{y}}\frac{\partial^{3}{f}}{\partial{y^{3}}}+2(\frac{\partial^{2}{f}}{\partial{x}})^2\frac{\partial{f}}{\partial{y}}\cdot\\& \frac{\partial^{3}{f}}{\partial{x^{2}}\partial{y}}\frac{\partial^{2}{f}}{\partial{y^{2}}}-(\frac{\partial{f}}{\partial{x}})^{2}\frac{\partial^{2}{f}}{\partial{x}\partial{y}}\frac{\partial{f}}{\partial{y}}\frac{\partial^{3}{f}}{\partial{x}\partial{y^{2}}}-(\frac{\partial{f}}{\partial{x}})^{2}\frac{\partial^{2}{f}}{\partial{x^{2}}}\frac{\partial{f}}{\partial{y}}\frac{\partial^{3}{f}}{\partial{y^{3}}}
\\&-\frac{\partial{f}}{\partial{x}}(\frac{\partial{f}}{\partial{y}})^{2}\frac{\partial^{3}{f}}{\partial{x^{3}}}\frac{\partial^{2}{f}}{\partial{y^{2}}}+2\frac{\partial{f}}{\partial{x}}\frac{\partial^{2}{f}}{\partial{x^{2}}}(\frac{\partial{f}}{\partial{y}})^{2}\frac{\partial^{3}{f}}{\partial{x}\partial{y^{2}}}-\frac{\partial{f}}{\partial{x}}\frac{\partial^{2}{f}}{\partial{x}\partial{y}}\cdot \\& (\frac{\partial{f}}{\partial{y}})^{2}\frac{\partial^{3}{f}}{\partial{x^{2}}\partial{y}}-\frac{\partial^{2}{f}}{\partial{x^{2}}}(\frac{\partial{f}}{\partial{y}})^{3}\frac{\partial^{3}{f}}{\partial{x^{2}}\partial{y}}+\frac{\partial^{2}{f}}{\partial{x}\partial{y}}(\frac{\partial{f}}{\partial{y}})^{3}\frac{\partial^{3}{f}}{\partial{x^{3}}}\}\\
\end{split}
\end{equation}
where $S$ is defined by
\begin{equation}\label{equ:34}
\begin{split}
S(x,y)=&(\frac{\partial{f}}{\partial{y}})^{3}\frac{\partial^{3}{f}}{\partial{x^{3}}}-3(\frac{\partial{f}}{\partial{y}})^{2}\frac{\partial{f}}{\partial{x}}\frac{\partial^{3}{f}}{\partial{x^{2}}\partial{y}}+3\frac{\partial{f}}{\partial{y}}(\frac{\partial{f}}{\partial{x}})^{2}\cdot \\&\frac{\partial^{3}{f}}{\partial{x}\partial{y^{2}}}
-(\frac{\partial{f}}{\partial{x}})^{3}\frac{\partial^{3}{f}}{\partial{y^{3}}}
\end{split}
\end{equation}

\hspace{-5mm}\textbf{Theorem 2.}
Let an image $f(x,y)$ be transformed by (\ref{equ:3}) into the image $g(u,v)$. Suppose that both $f(x,y)$ and $g(u,v)$ have the third-order partial derivatives, then we have
\begin{equation}\label{equ:35}
  RPDI_{2}^{'}(u,v)= \frac{RPDI_{2}(x,y)}{J^{6}}
\end{equation}
where $RPDI_{2}^{'}(u,v)$ is defined in the way similar to that of $RPDI_{1}^{'}(u,v)$. Theorem 2 can be proved very easily by $Maple$, too.

We believe that there must be other structural formulas of $RPDIs$. Especially, \cite{7} proposed that the isomorphism between differential invariants and geometric moment invariants under general affine transformations. This makes it very easy to obtain affine differential invariants. As we all know, the affine transformation group is a subgroup of the projective transformation group. A projective invariant must be invariant to affine transformations. Thus, while affine differential invariants are well established, $RPDIs$ could be screened out from affine ones.
\section{The Structural framework of PIs}\label{sec:5}
In this section, we will present how to construct $PIs$, firstly. Then, a new weighted moment, which can be used to calculate $PIs$, will be defined. Finally, some instances of $PIs$ will be given for experiments in Section~\ref{sec:6}.
\subsection{The Construction of PIs}
By using (\ref{equ:12}), (\ref{equ:20}), (\ref{equ:27}) and (\ref{equ:33}), we can obtain two construction methods of $PIs$.

\hspace{-5mm}\textbf{Theorem 3.}
Let an image $f(x,y)$ be transformed by (\ref{equ:3}) into the image $g(u,v)$. $(u_{1},v_{1})$, $(u_{2},v_{2})$ and $(u_{3},v_{3})$ in $g(u,v)$ are the corresponding points of $(x_{1},y_{1})$, $(x_{2},y_{2})$ and $(x_{3},y_{3})$. Suppose that $f(x,y)$ and $g(u,v)$ both have the second-order partial derivatives. We have
\begin{equation}\label{equ:36}
\begin{split}
PI_{1}^{n}(f)&=\int...\int (d(1,2,3)D(1,2,3))^{n}\cdot\\&\sqrt{RPDI_{1}(x_{1},y_{1})}\sqrt{RPDI_{1}(x_{2},y_{2})}\cdot \\&\sqrt{RPDI_{1}(x_{3},y_{3})}f(x_{1},y_{1})f(x_{2},y_{2})f(x_{3},y_{3})\\&dx_{1}dy_{1}dx_{2}dy_{2}dx_{3}dy_{3}
\end{split}
\end{equation}
where $n=0,1,2,...$. Then, there is a relation
\begin{equation}\label{equ:37}
  PI_{1}^{n}(g)=PI_{1}^{n}(f)
\end{equation}
where
\begin{equation}\label{equ:38}
\begin{split}
PI_{1}^{n}(g)&=\int...\int (d^{'}(1,2,3)D^{'}(1,2,3))^{n}\cdot\\&\sqrt{RPDI_{1}^{'}(u_{1},v_{1})}\sqrt{RPDI_{1}^{'}(u_{2},v_{2})}\cdot \\&\sqrt{RPDI_{1}^{'}(u_{3},v_{3})}g(u_{1},v_{1})g(u_{2},v_{2})g(u_{3},v_{3})\\&du_{1}dv_{1}du_{2}dv_{2}du_{3}dv_{3}
\end{split}
\end{equation}
\textbf{Proof.} According to (\ref{equ:5}), (\ref{equ:14}), (\ref{equ:22}) and (\ref{equ:28}) in Section~\ref{sec:2.2}, Section~\ref{sec:3.2} and Section~\ref{sec:4}, we have
\begin{equation}\label{equ:39}
\begin{split}
PI_{1}^{n}(g)&=\int...\int (\sqrt[3]{J(x_{1},y_{1})}\sqrt[3]{J(x_{2},y_{2})}\sqrt[3]{J(x_{3},y_{3})}\cdot\\&d(1,2,3)\frac{1}{\sqrt[3]{J(x_{1},y_{1})}}\frac{1}{\sqrt[3]{J(x_{2},y_{2})}}\frac{1}{\sqrt[3]{J(x_{3},y_{3})}}\cdot\\&D(1,2,3))^{n} \frac{\sqrt{RPDI_{1}(x_{1},y_{1})}}{J(x_{1},y_{1})}\frac{\sqrt{RPDI_{1}(x_{2},y_{2})}}{J(x_{2},y_{2})}\cdot \\&\frac{\sqrt{RPDI_{1}(x_{3},y_{3})}}{J(x_{3},y_{3})}f(x_{1},y_{1})f(x_{2},y_{2})f(x_{3},y_{3})\cdot\\&J(x_{1},y_{1})J(x_{2},y_{2})J(x_{3},y_{3})dx_{1}dy_{1}dx_{2}dy_{2}dx_{3}dy_{3}
\\&=PI_{1}^{n}(f)
\end{split}
\end{equation}

Therefore, it is proved that $PI_{1}^{n}(f)$ has invariance to general projective transformations. Similarly, we can use $RPDI_{2}$ to construct $PIs$.

\hspace{-5mm}\textbf{Theorem 4.}~Let an image $f(x,y)$ be transformed by (\ref{equ:3}) into the image $g(u,v)$. $(u_{1},v_{1})$, $(u_{2},v_{2})$ and $(u_{3},v_{3})$ in $g(u,v)$ are the corresponding points of $(x_{1},y_{1})$, $(x_{2},y_{2})$ and $(x_{3},y_{3})$. Suppose that $f(x,y)$ and $g(u,v)$ both have the third-order partial derivatives. We have
\begin{equation}\label{equ:40}
\begin{split}
PI_{2}^{n}(f)&=\int...\int (d(1,2,3)D(1,2,3))^{n}\cdot\\&\sqrt[6]{RPDI_{2}(x_{1},y_{1})}\sqrt[6]{RPDI_{2}(x_{2},y_{2})}\cdot \\&\sqrt[6]{RPDI_{2}(x_{3},y_{3})}f(x_{1},y_{1})f(x_{2},y_{2})f(x_{3},y_{3})\\&dx_{1}dy_{1}dx_{2}dy_{2}dx_{3}dy_{3}
\end{split}
\end{equation}
where $n=0,1,2,...$. Then, there is a relation
\begin{equation}\label{equ:41}
  PI_{2}^{n}(g)=PI_{2}^{n}(f)
\end{equation}
where
\begin{equation}\label{equ:42}
\begin{split}
PI_{2}^{n}(g)&=\int...\int (d^{'}(1,2,3)D^{'}(1,2,3))^{n}\cdot\\&\sqrt[6]{RPDI_{2}^{'}(u_{1},v_{1})}\sqrt[6]{RPDI_{2}^{'}(u_{2},v_{2})}\cdot \\&\sqrt[6]{RPDI_{2}^{'}(u_{3},v_{3})}g(u_{1},v_{1})g(u_{2},v_{2})g(u_{3},v_{3})\\&du_{1}dv_{1}du_{2}dv_{2}du_{3}dv_{3}
\end{split}
\end{equation}

The proof of Theorem 4 is similar to that of Theorem 3, by using (\ref{equ:5}), (\ref{equ:14}), (\ref{equ:22}) and (\ref{equ:35}). Firstly, we must point out that there are many other structural formulas which can be designed. Here is not listed one by one. Then, it is obviously that for any n, $PI_{1}^{n}$ and $PI_{2}^{n}$ are not always zero. Because $PI_{1}^{n}=(-1)^{2n}PI_{1}^{n}=PI_{1}^{n}$ and $PI_{2}^{n}=(-1)^{2n}PI_{2}^{n}=PI_{2}^{n}$. Thus, we don't make the mistake in \cite{17, 25}.

\subsection{The Definition of Projective Weighted Moment}
In order to calculate invariants more conveniently, $PIs$ have to be represented as polynomials of some kinds of moments. So, we need to give the definitions of these moments.

\hspace{-5mm}\textbf{Definition 3.}
For an image $f(x,y)$, we define two kinds of weighted moments, which are named projective weighted moments ($PMs$).
\begin{equation}\label{equ:43}
\begin{split}
PM^{1}_{pqrst}=& \iint x^{p}y^{q}(\frac{\partial{f}}{\partial{x}})^{r}(\frac{\partial{f}}{\partial{y}})^{s}(x\frac{\partial{f}}{\partial{x}}+y\frac{\partial{f}}{\partial{y}})^{t}\cdot \\&\sqrt{RPDI_{1}(x,y)}f(x,y)dxdy
\end{split}
\end{equation}
\begin{equation}\label{equ:44}
\begin{split}
PM^{2}_{pqrst}=& \iint x^{p}y^{q}(\frac{\partial{f}}{\partial{x}})^{r}(\frac{\partial{f}}{\partial{y}})^{s}(x\frac{\partial{f}}{\partial{x}}+y\frac{\partial{f}}{\partial{y}})^{t}\cdot \\&\sqrt[6]{RPDI_{2}(x,y)}f(x,y)dxdy
\end{split}
\end{equation}

Comparing with (\ref{equ:26}), it can be found that $RPDIs$ are used to construct the weight functions.
\subsection{The Instances of PIs}
We can use (\ref{equ:36}) and (\ref{equ:40}) to construct instances of $PIs$. By setting $n=0,1$, we can get 4 instances. Their expanded forms are defined by (\ref{equ:45}), (\ref{equ:46}) and (\ref{equ:47}).
\begin{equation}\label{equ:45}
\begin{split}
PI_{1}^{0}=(PM^{1}_{00000})^{3}~~~~~~~~PI_{2}^{0}=(PM^{2}_{00000})^{3}
\end{split}
\end{equation}
\begin{equation}\label{equ:46}
\begin{split}
PI_{1}^{1}=&6PM^{1}_{00001}PM^{1}_{01010}PM^{1}_{10100}-6PM^{1}_{00001}PM^{1}_{01100}\cdot\\&PM^{1}_{10010}-6PM^{1}_{00010}PM^{1}_{01001}PM^{1}_{10100}
+6PM^{1}_{00010}\cdot\\&PM^{1}_{01100}PM^{1}_{10001}+6PM^{1}_{00100}PM^{1}_{01001}PM^{1}_{10010}\cdot\\&
-6PM^{1}_{00100}PM^{1}_{01010}PM^{1}_{10001}
\end{split}
\end{equation}
\begin{equation}\label{equ:47}
\begin{split}
PI_{2}^{1}=&6PM^{2}_{00001}PM^{2}_{01010}PM^{2}_{10100}-6PM^{2}_{00001}PM^{2}_{01100}\cdot\\&PM^{2}_{10010}-6PM^{2}_{00010}PM^{2}_{01001}PM^{2}_{10100}
+6PM^{2}_{00010}\cdot\\&PM^{2}_{01100}PM^{2}_{10001}+6PM^{2}_{00100}PM^{2}_{01001}PM^{2}_{10010}\cdot\\&
-6PM^{2}_{00100}PM^{2}_{01010}PM^{2}_{10001}
\end{split}
\end{equation}

Obviously, compared with \cite{17, 25}, these invariants are made up of finite terms and are not always zero. But we should also point out that these projective invariants can be applied based on the hypothesis that we are able to calculate the partial derivatives accurately on discrete images. We will discuss this issue in detail in the Section \ref{sec:6}.

\section{Experiments and discussions}\label{sec:6}

In this section, we design some experiments to evaluate the performance of $PIs$. Firstly, the possible sources of calculation error are analyzed in detail. According to these analyzing results, we design some methods to reduce the calculation error of $PIs$, which make $PIs$ have better stability and discriminability. Then, retrieval experiment is performed on the dataset consisting of synthetic images. For comparison, we choose 4 commonly used moment invariants. Finally, image retrieval and classification based on real datasets are carried out. Various experimental results show that $PIs$ do have better properties for general projective transformations than other traditional moment invariants.

\subsection{The Error Sources Analysis}
The premise that invariants can be applied is that their numerical values have good stability and discriminability. Therefore, it is necessary to reduce the calculation error as much as possible. By observing the construction of $PIs$, we can find that the error may mainly come from two aspects.

\begin{figure*}
  \centering
  \subfloat[$D_{f}$ and $D_{g}$ are equivalent, when $f(x,y)$ and $g(u,v)$ are continuous.]{\includegraphics[height=4cm,width=8.0cm]{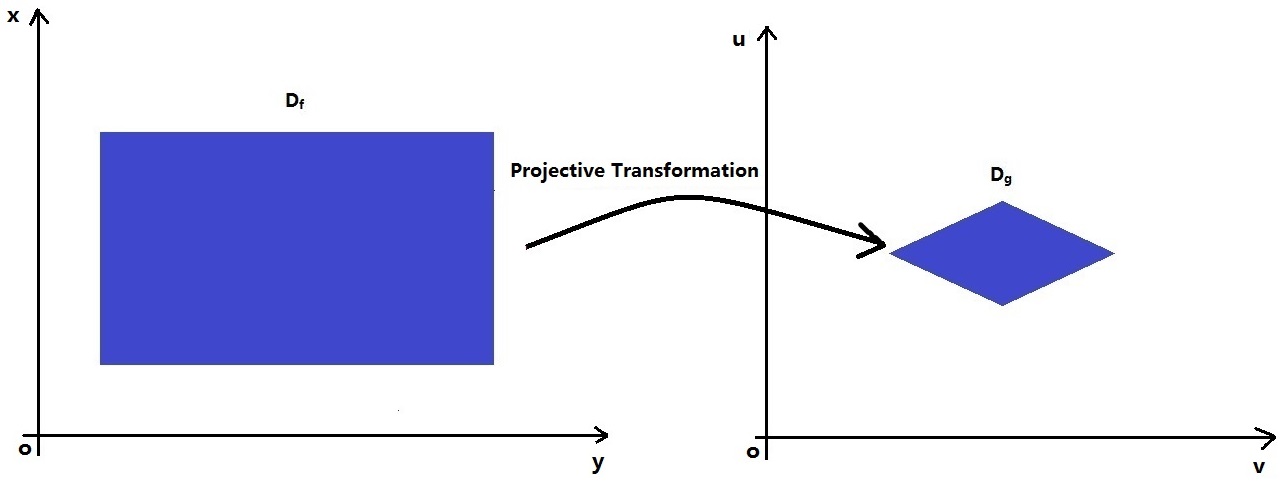}\label{Fig:2(a)}\hfill}~~~~~~~~
  \subfloat[The number of pixels contained in two images is not equal, when $f(x,y)$ and $g(u,v)$ are discrete images.]{\includegraphics[height=4cm,width=8.0cm]{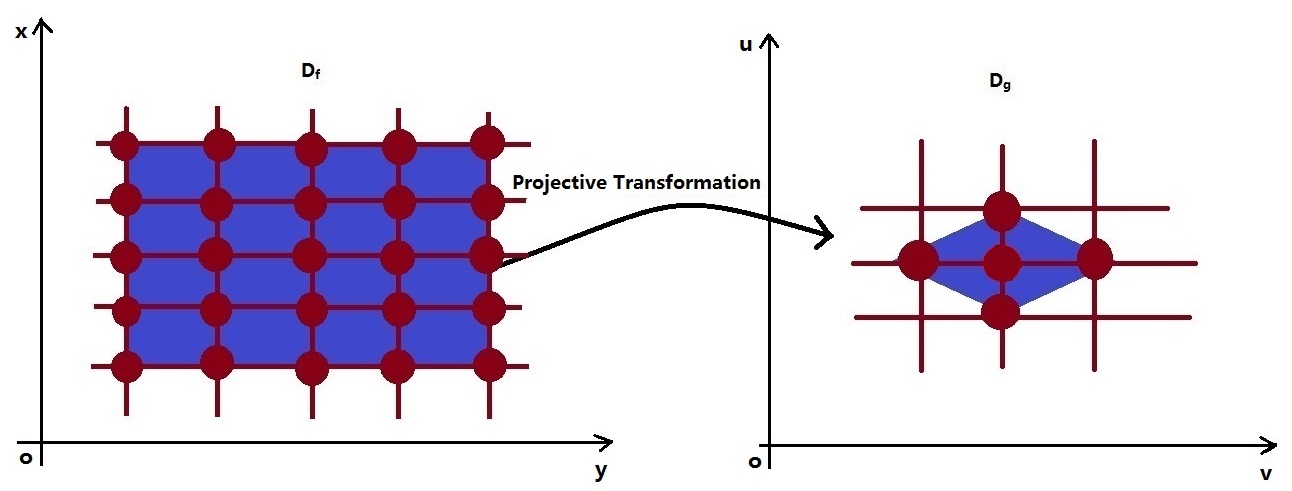}\label{Fig:2(b)}\hfill}
  \caption{The sketch map showing the change in the number of elements in the domain, when the function $f(x,y)$ is transformed into $g(u,v)$    by a projective transformation.}
\end{figure*}

\begin{enumerate}
  \item In Section~\ref{sec:4} and \ref{sec:5}, we suppose that the image $f(x,y)$ is a continuous function with the second-order or third-order partial derivatives. In practice, however, $f(x,y)$ is a discrete function. So, we can not directly obtain the exact values of these partial derivatives, and can only use other methods to approach. How to estimate the values of partial derivatives will greatly affect the performance of $PIs$.
  \item When $f(x,y)$ is a continuous function, we define its domain $D_{f}\subset R^{2}$. Suppose that $f(x,y)$ is transformed into $g(u,v)$ by (\ref{equ:3}). The domain of $g(u,v)$ is $D_{g}\subset R^{2}$. We need to point out that both $D_{f}$ and $D_{g}$ are uncountable sets. Thus, they are equivalent which means they can be placed in one-to-one correspondence, which are shown in Fig.~\ref{Fig:2(a)}. But when $f(x,y)$ and $g(u,v)$ are discrete images, $D_{f}$ and $D_{g}$ are finite sets. The equivalent relation between $D_{f}$ and $D_{g}$ doesn't exist. As shown in Fig.~\ref{Fig:2(b)}, the number of pixels contained in two images is not equal. In \cite{17,25}, the authors didn't find this problem. We think that it is necessary to reduce the error caused by the change in the number of pixels.
\end{enumerate}

To solve the first problem, we employ the derivatives of the Gaussian function as filters to compute the derivatives of an image function via convolution. Many researches have shown that this method can achieve good results \cite{15,16}. The 2D zero-mean Gaussian function and its partial derivatives are defined by
\begin{equation}\label{equ:48}
\begin{split}
  &G(x,y)=\frac{1}{2\pi\sigma^{2}}\emph{e}^{-\frac{x^{2}+y^{2}}{2\sigma^{2}}}~~~~~~~~~
  \frac{\partial{G}}{\partial{x}}=-\frac{x}{\sigma^{2}}G\\
  &\frac{\partial{G}}{\partial{y}}=-\frac{y}{\sigma^{2}}G~~~~~~~~~~~~~~~~~~~~~~~~~
  \frac{\partial^{2}{G}}{\partial{x^{2}}}=\frac{(x^{2}-\sigma^{2})}{\sigma^{4}}G\\
  &\frac{\partial^{2}{G}}{\partial{x}\partial{y}}=\frac{xy}{\sigma^{4}}G~~~~~~~~~~~~~~~~~~~~~~~~
  \frac{\partial^{2}{G}}{\partial{y^{2}}}=\frac{(y^{2}-\sigma^{2})}{\sigma^{4}}G\\
  &\frac{\partial^{3}{G}}{\partial{x^{3}}}=\frac{(3\sigma^{2}x-x^{3})}{\sigma^{6}}G~~~~~~~~~~
  \frac{\partial^{3}{G}}{\partial{x^{2}\partial{y}}}=\frac{(\sigma^{2}y-x^{2}y)}{\sigma^{6}}G\\
  &\frac{\partial^{3}{G}}{\partial{x}\partial{y^{2}}}=\frac{(\sigma^{2}x-xy^{2})}{\sigma^{6}}G~~~~~~~
  \frac{\partial^{3}{G}}{\partial{y^{3}}}=\frac{(3\sigma^{2}y-y^{3})}{\sigma^{6}}G
\end{split}
\end{equation}
where $\sigma$ is the standard deviation.

By using (\ref{equ:48}) to convolved with the image function $f(x,y)$, partial derivatives of $f(x,y)$ can be obtained. For example,
\begin{equation}\label{equ:49}
 \frac{\partial{f}}{\partial{x}}=\frac{\partial{G}}{\partial{x}}\circledast f(x,y)
\end{equation}
where $\circledast$ represents the convolution operation. For discrete images, we use (\ref{equ:48}) to convolve with the $N \times N$ neighborhood of $(x_{i},y_{j})$. In general, $N$ is odd and $\sigma=\frac{N-1}{6}$.

For comparison, we also choose two other methods, the least squares method and the weighted least squares method \cite{4}. A general polynomial surface of the third-order is given by
\begin{equation}\label{equ:50}
\begin{split}
  &Z_{i,j}(x,y)=a_{0}+a_{1}(x-x_{i})+a_{2}(y-y_{j})+a_{3}(x-x_{i})^{2}+\\
  &a_{4}(y-y_{j})^{2}+a_{5}(x-x_{i})(y-y_{j})+a_{6}(x-x_{i})^{3}+
  a_{7}(y-y_{j})^{3}\\&+a_{8}(x-x_{i})^{2}(y-y_{j})+a_{9}(x-x_{i})(y-y_{j})^{2}
\end{split}
\end{equation}

By using (\ref{equ:50}) and the least squares method, we can fit out the surface in the $N \times N$ neighborhood of $(x_{i},y_{j})$. Also, $N$ is odd. So, the values of $(a_{0}, a_{1}, a_{2},...,a_{9})$ can be obtained, which can be used to calculate the partial derivatives in the point $(x_{i},y_{j})$. For example,
\begin{equation}\label{equ:51}
 \frac{\partial{f}}{\partial{x}}\bigg |_{(x_{i},y_{j})}=a_{1}
\end{equation}

The weighted least squares method is similar to the least square method. But it sets the weight for each point $(x,y)$ in the $N\times N$ neighborhood of $(x_{i}, y_{j})$ according to the distance between $(x,y)$ and $(x_{i},y_{j})$. For details of the weighted least squares method, you can see \cite{4}. In our paper, the weight is defined by
\begin{equation}\label{equ:52}
  W_{i,j}(x,y)=\frac{1+\frac{N-1}{2}\sqrt{\frac{N-1}{2}}-\sqrt{(x-x_{i})^{2}+(y-y_{j})^{2}}}{\frac{N-1}{2}\sqrt{\frac{N-1}{2}}}
\end{equation}

To solve the second problem, the number of pixels is normalized. As we all known, the Fourier transform of a continuous function $f(x,y)$ is defined by
\begin{equation}\label{equ:53}
  F(u,v)=\iint f(x,y)e^{-j2\pi(ux+vy)}dxdy
\end{equation}

But when $f(x,y)$ is a discrete image, the Fourier transform is defined by
\begin{equation}\label{equ:54}
  F(u,v)=\frac{1}{M\times N}\sum^{M-1}_{x=0}\sum^{N-1}_{y=0}f(x,y)e^{-j2\pi(\frac{ux}{M}+\frac{vy}{N})}
\end{equation}
where $M \times N$ is the size of the image. Obviously, $\frac{1}{M\times N}$ is used to normalized the number of pixels.

Similarly, when $f(x,y)$ is a discrete image, $PMs$ should be normalized by the number of pixels, which means
\begin{equation}\label{equ:55}
\begin{split}
   PM^{1}_{pqrst}=&\frac{1}{Num}\sum_{(x,y)\in D_{f}} x^{p}y^{q}(\frac{\partial{f}}{\partial{x}})^{r}(\frac{\partial{f}}{\partial{y}})^{s}(x\frac{\partial{f}}{\partial{x}}+y\frac{\partial{f}}{\partial{y}})^{t}\cdot \\& \sqrt{RPDI_{1}(x,y)}f(x,y)
\end{split}
\end{equation}
\begin{equation}\label{equ:56}
\begin{split}
   PM^{2}_{pqrst}=&\frac{1}{Num}\sum_{(x,y)\in D_{f}} x^{p}y^{q}(\frac{\partial{f}}{\partial{x}})^{r}(\frac{\partial{f}}{\partial{y}})^{s}(x\frac{\partial{f}}{\partial{x}}+y\frac{\partial{f}}{\partial{y}})^{t}\cdot \\& \sqrt[6]{RPDI_{2}(x,y)}f(x,y)
\end{split}
\end{equation}
where $D_{f}$ is the domain of $f(x,y)$, $Num$ represents the number of nonzero pixels in the image $f(x,y)$. However, the previous work \cite{17,25} ignored this problem. It may lead to huge error in the calculation of $PIs$, when the image is scaled.

\subsection{Numerical Verification of PIs}
Now, we verify that the methods designed to reduce the calculation error are valid. Firstly, we choose 5 images from the USC-SIPI (http://sipi.usc.edu/database/). Each image is transformed by 5 general projective transformations which are defined in Table.~\ref{Tab:1}.
\begin{table}[h]
\caption{Five general projective transformations (r=1)}
\centering
\begin{tabular}{cccccccccc}
  \toprule[1.1pt]
  No. & a & b & c & d & e & f & p & q \\
  \toprule[1.1pt]
  1 & 3.0 & -0.2 & -20 & -0.1 & 1.3 & 300 & 0.006 & -0.0001  \\
  \hline
  2 & -1.9 & 0.02 & 40 & 0.1 & -1.8 & 50 & 0.002 & 0.002 \\
  \hline
  3 & 1.3 & 0.2 & 45 & 0.1 & 1.6 & -30 & 0.002 & 0.002  \\
  \hline
  4 & 1.6 & 0.4 & -75 & 0.2 & 1.7 & -80 & 0.001 & 0.002  \\
  \hline
  5 & 0.7 & 0.2 & -55 & -0.1 & 1.3 & 45 & -0.0001 & -0.0001  \\
  \toprule[1.1pt]
\end{tabular}
\label{Tab:1}
\end{table}

\begin{figure}
  \centering
  \includegraphics[height=9.0cm,width=8.0cm]{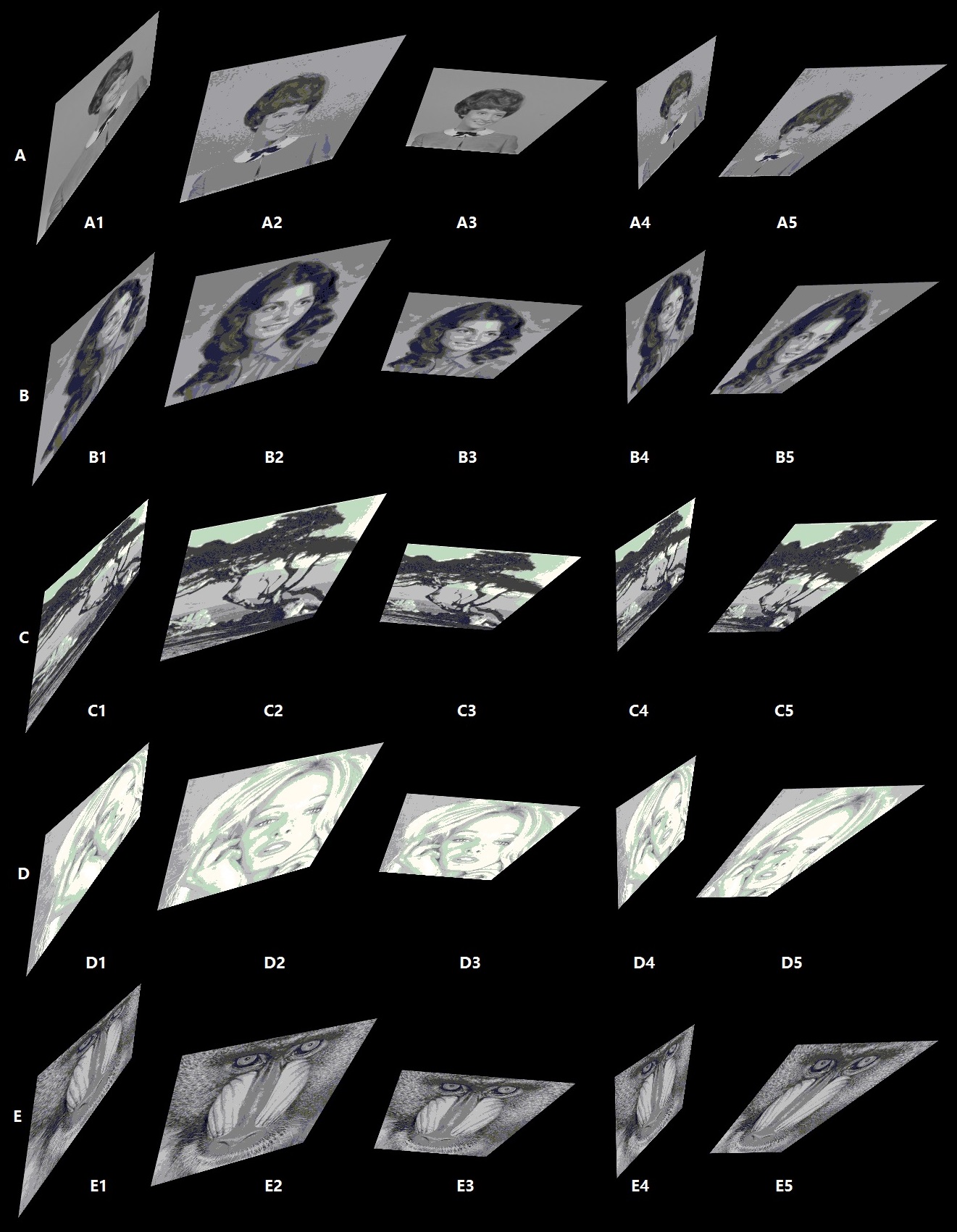}\\
  \caption{Test images, each image is transformed by 5 general projective transformations.}\label{Fig:3}
\end{figure}

Thus, 25 images are obtained, which are shown in Fig~\ref{Fig:3}. They can be divided into 5 groups, each group contains 5 images.

For comparison, we use the Gaussian derivative method, the least squares method and the weighted least squares method to compute $(PI^{0}_{1}, PI^{0}_{2},PI^{1}_{1}, PI^{2}_{1})$. The neighborhood size of a point in an image is set to $5 \times 5$, $7 \times 7$, $9 \times 9$, $11 \times 11$, $13 \times 13$, $15 \times 15$ and $17 \times 17$, respectively. In all three methods, the number of pixels is normalized. In order to evaluate the calculation results, we use the average relative error $(ARE)$, which is defined by
\begin{equation}\label{equ:57}
  ARE=\frac{1}{5}\sum_{X}Error(X)
\end{equation}
where
\begin{equation}\label{equ:58}
   Error(X)=\frac{\max\limits_{j}\{X_{j}\}-\min\limits_{k}\{X_{k}\}}{|\max\limits_{j}\{X_{j}\}|+|\min\limits_{k}\{X_{k}\}|}
\end{equation}
$X\in\{A,B,C,D,E\}$, $j,k\in\{1,2,3,4,5\}$.

\begin{figure*}
  \centering
   \subfloat[The ARE of $PI^{0}_{1}$]{\includegraphics[height=4cm,width=7.3cm]{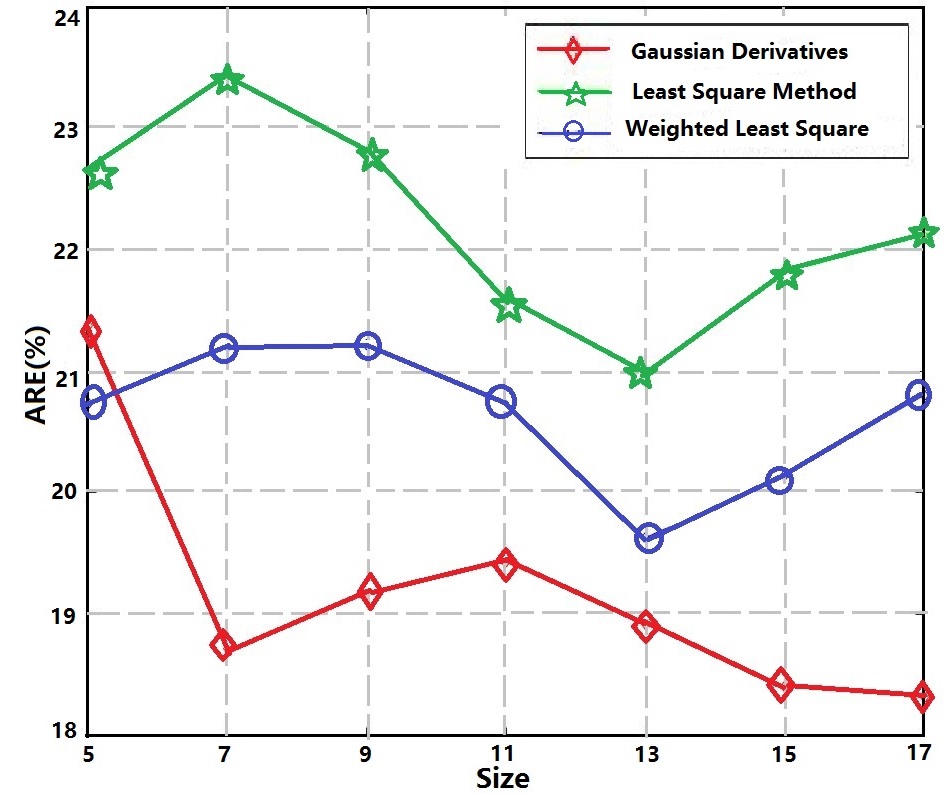}\hfill}~~~~~
   \subfloat[The ARE of $PI^{0}_{2}$]{\includegraphics[height=4cm,width=7.3cm]{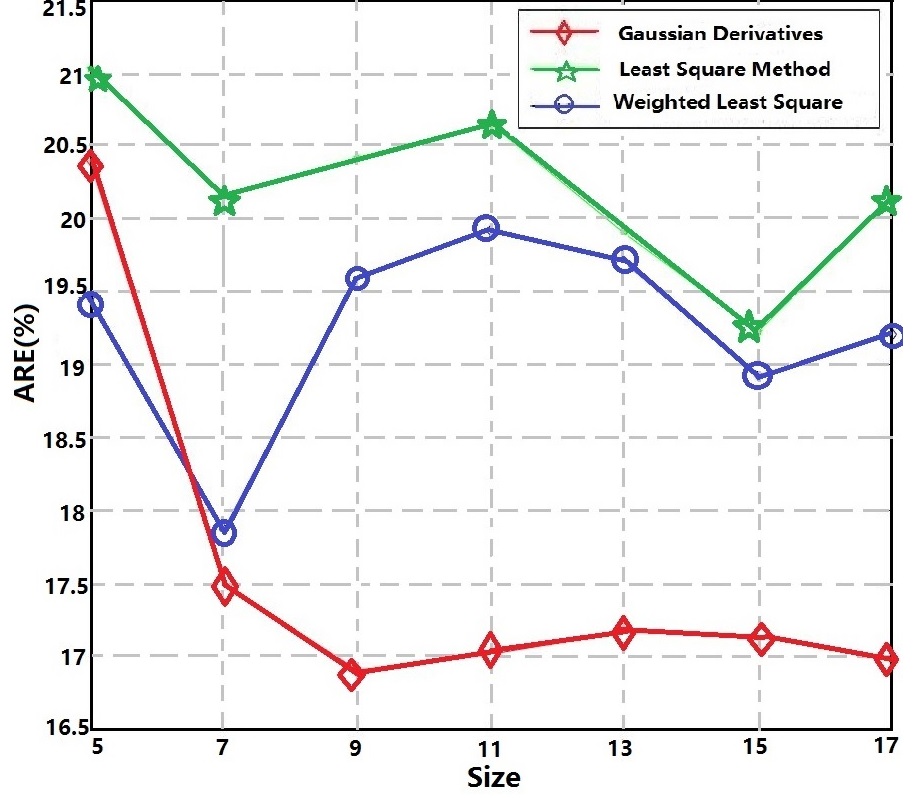}}\\
  \subfloat[The ARE of $PI^{1}_{1}$]{\includegraphics[height=4cm,width=7.3cm]{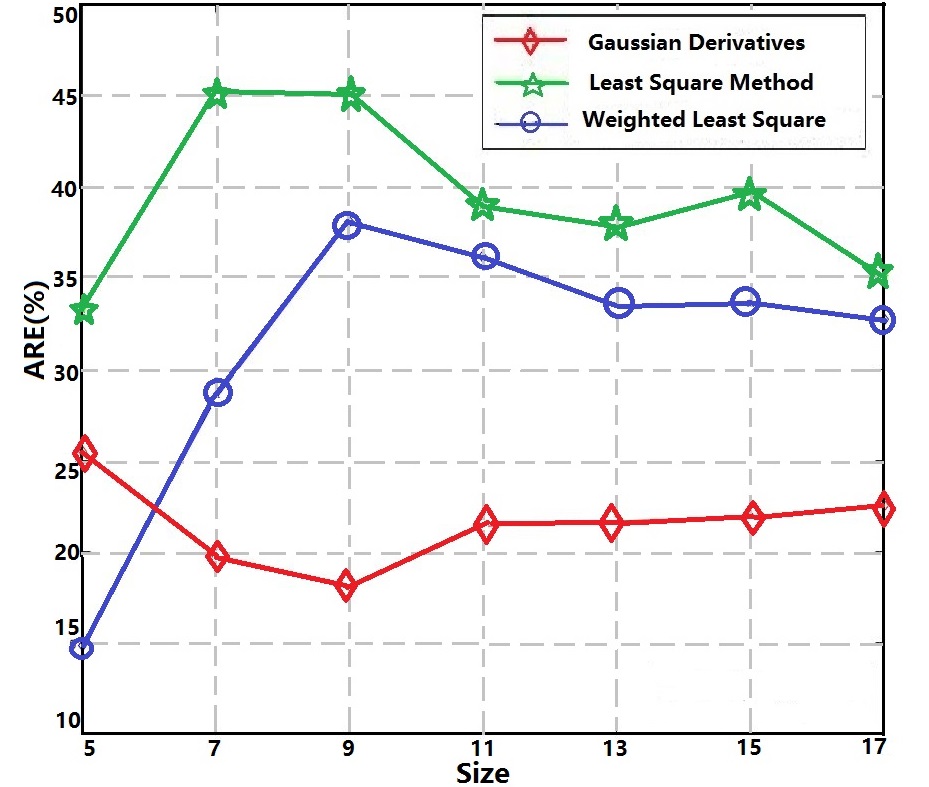}\hfill}~~~~~
   \subfloat[The ARE of $PI^{1}_{2}$]{\includegraphics[height=4cm,width=7.3cm]{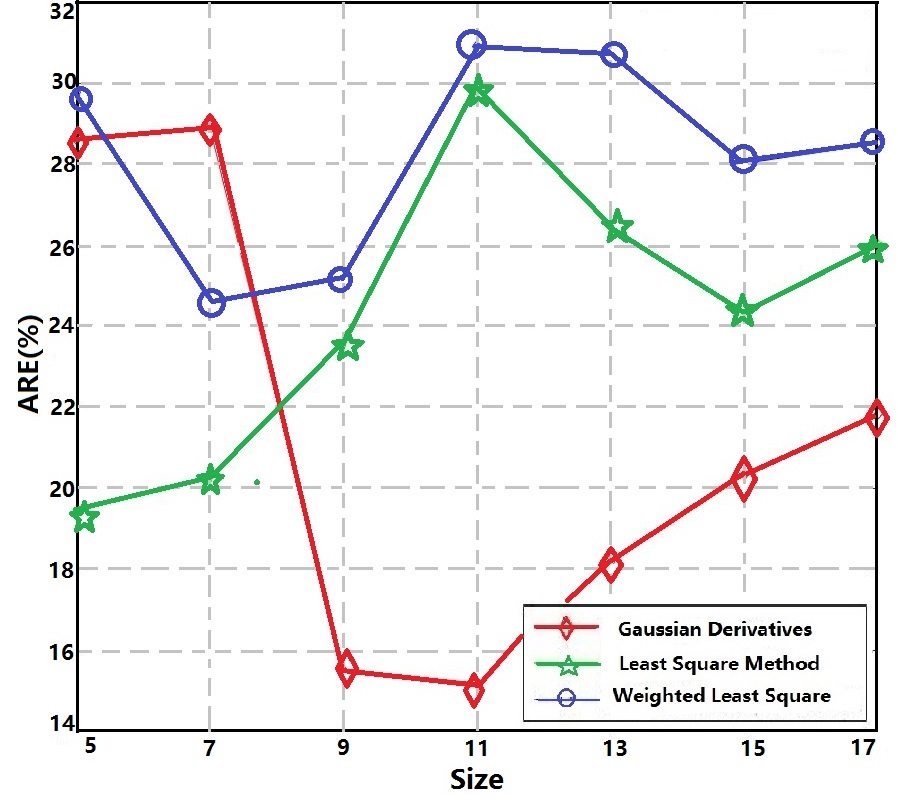}}\\
  \caption{The ARE of $PIs$ by using different calculation methods of partial derivatives and different neighborhood sizes. The red, green and blue curves are obtained by using the Gaussian derivatives, the weighted least squares method and the least squares method, respectively.}\label{Fig:4}
\end{figure*}

According to Fig~\ref{Fig:4}, we can find that the Gaussian derivative method has better properties for $PIs$ than other methods. In most cases, the weighted least squares method is better than the least squares method. With the increase of the neighborhood size, the error of $PIs$ is reduced. But this trend is not always. Considering the computational stability and time cost, we decide to use the Gaussian derivative method and set the neighborhood size to $9 \times 9$ $(\sigma=1.33)$ in next experiments. Based on this setting, the numerical values of $PIs$ of 25 test images are shown in Table.~\ref{Tab:2}. The error in Table.~\ref{Tab:2} is defined by (\ref{equ:58}).

\begin{table}
\caption{The numerical values of $PIs$ of 25 test images.}
\centering
\begin{tabular}{p{3.3mm}<{\centering}p{13.2mm}<{\centering}p{13.2mm}<{\centering}p{13.2mm}<{\centering}p{13.5mm}<{\centering}}
  \toprule[1.1pt]
  No. & $PI^{0}_{1}$ & $PI^{0}_{2}$ & $PI^{1}_{1}$ & $PI^{1}_{2}$\\
  \toprule[1.1pt]
  A1 & $7.8\cdot10^{-3}$ & $1.6\cdot10^{-2}$ & $1.7\cdot10^{3}$ & $1.1\cdot10^{4}$ \\
  A2 & $5.4\cdot10^{-3}$ & $1.2\cdot10^{-2}$ & $1.0\cdot10^{3}$ & $1.1\cdot10^{4}$ \\
  A3 & $6.7\cdot10^{-3}$ & $1.4\cdot10^{-2}$ & $1.2\cdot10^{3}$ & $9.1\cdot10^{3}$ \\
  A4 & $8.8\cdot10^{-3}$ & $1.8\cdot10^{-2}$ & $1.4\cdot10^{3}$ & $8.8\cdot10^{3}$ \\
  A5 & $6.9\cdot10^{-3}$ & $1.5\cdot10^{-2}$ & $1.3\cdot10^{3}$ & $8.6\cdot10^{3}$ \\
  \hline
  Error & $23.93\%$ & $19.96\%$ & $23.86\%$ & $15.07\%$\\
  \hline
  B1 & $2.5\cdot 10^{-2}$ & $4.9\cdot 10^{-2}$ & $1.3\cdot 10^{4}$ & $5.2\cdot 10^{-21}$ \\
  B2 & $2.2\cdot 10^{-2}$ & $4.2\cdot 10^{-2}$ & $1.2\cdot 10^{4}$ & $4.1\cdot 10^{-21}$ \\
  B3 & $2.6\cdot 10^{-2}$ & $4.9\cdot 10^{-2}$ & $1.5\cdot 10^{4}$ & $3.6\cdot 10^{-21}$ \\
  B4 & $3.0\cdot 10^{-2}$ & $5.8\cdot 10^{-2}$ & $1.5\cdot 10^{4}$ & $3.9\cdot 10^{-21}$ \\
  B5 & $2.4\cdot 10^{-2}$ & $4.6\cdot 10^{-2}$ & $1.4\cdot 10^{4}$ & $4.0\cdot 10^{-21}$ \\
  \hline
  Error & $16.25\%$ & $15.77\%$ & $11.16\%$ & $19.44\%$\\
  \hline
  C1 & $1.3\cdot 10^{-2}$ & $2.8\cdot 10^{-2}$ & $6.4\cdot 10^{4}$ & $4.1\cdot 10^{5}$ \\
  C2 & $9.9\cdot 10^{-3}$ & $2.1\cdot 10^{-2}$ & $4.9\cdot 10^{4}$ & $4.0\cdot 10^{5}$ \\
  C3 & $1.2\cdot 10^{-2}$ & $2.5\cdot 10^{-2}$ & $5.7\cdot 10^{4}$ & $3.6\cdot 10^{5}$ \\
  C4 & $1.5\cdot 10^{-2}$ & $3.0\cdot 10^{-2}$ & $5.8\cdot 10^{4}$ & $3.4\cdot 10^{5}$ \\
  C5 & $1.2\cdot 10^{-2}$ & $2.6\cdot 10^{-2}$ & $5.7\cdot 10^{4}$ & $3.4\cdot 10^{5}$ \\
  \hline
  Error & $20.00\%$ & $17.02\%$ & $13.60\%$ & $9.70\%$\\
  \hline
  D1 & $1.5\cdot 10^{-2}$ & $3.2\cdot 10^{-2}$ & $2.4\cdot 10^{3}$ & $1.9\cdot 10^{4}$ \\
  D2 & $1.4\cdot 10^{-2}$ & $3.1\cdot 10^{-2}$ & $2.5\cdot 10^{3}$ & $2.2\cdot 10^{4}$ \\
  D3 & $1.5\cdot 10^{-2}$ & $3.4\cdot 10^{-2}$ & $2.6\cdot 10^{3}$ & $2.0\cdot 10^{4}$ \\
  D4 & $1.7\cdot 10^{-2}$ & $3.6\cdot 10^{-2}$ & $2.2\cdot 10^{3}$ & $1.6\cdot 10^{4}$ \\
  D5 & $1.6\cdot 10^{-2}$ & $3.5\cdot 10^{-2}$ & $2.2\cdot 10^{3}$ & $1.7\cdot 10^{4}$ \\
  \hline
  Error & $8.98\%$ & $8.06\%$ & $9.30\%$ & $14.65\%$\\
  \hline
  E1 & $1.1\cdot 10^{-2}$ & $2.5\cdot 10^{-2}$ & $8.6\cdot 10^{2}$ & $6.1\cdot 10^{3}$ \\
  E2 & $8.3\cdot 10^{-3}$ & $1.8\cdot 10^{-2}$ & $4.4\cdot 10^{2}$ & $4.7\cdot 10^{3}$ \\
  E3 & $1.1\cdot 10^{-2}$ & $2.3\cdot 10^{-2}$ & $5.4\cdot 10^{2}$ & $4.2\cdot 10^{3}$ \\
  E4 & $1.4\cdot 10^{-2}$ & $3.0\cdot 10^{-2}$ & $6.3\cdot 10^{2}$ & $4.5\cdot 10^{3}$ \\
  E5 & $1.1\cdot 10^{-2}$ & $2.4\cdot 10^{-2}$ & $6.7\cdot 10^{2}$ & $4.7\cdot 10^{3}$ \\
  \hline
  Error & $26.70\%$ & $23.73\%$ & $32.93\%$ & $18.77\%$\\
  \toprule[1.1pt]
\end{tabular}
\label{Tab:2}
\end{table}

\begin{figure}
  \centering
  \includegraphics[height=6.5cm,width=6.8cm]{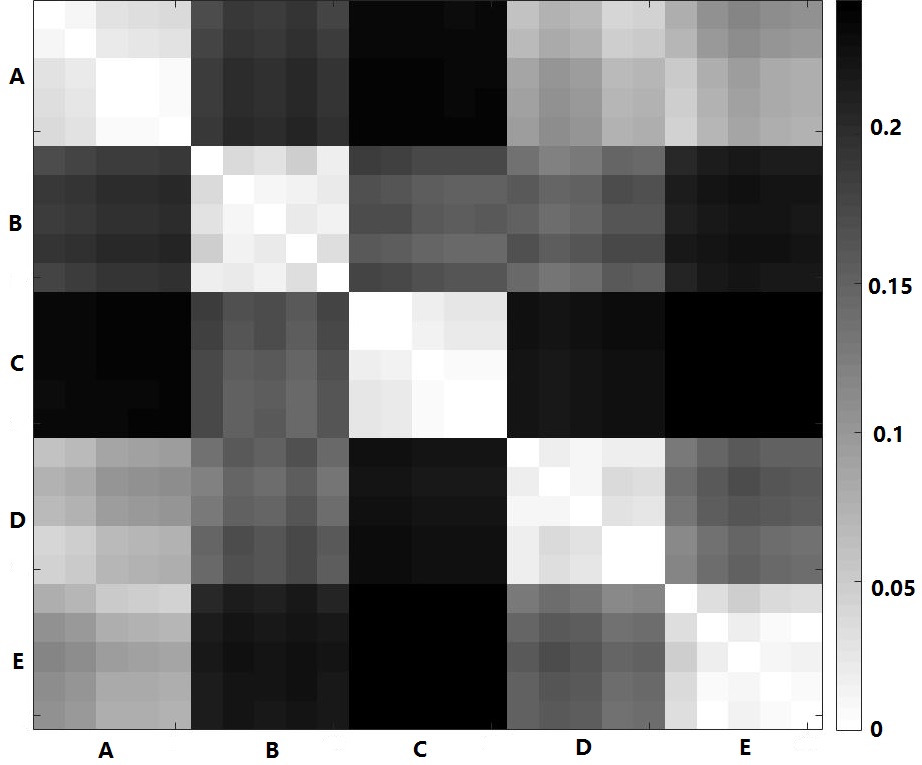}\\
  \caption{The visualization of the distance matrix. As the distance increases, the color changes from white to black.}\label{Fig:5}
\end{figure}
\begin{figure}
  \centering
  \includegraphics[height=6.5cm,width=6.8cm]{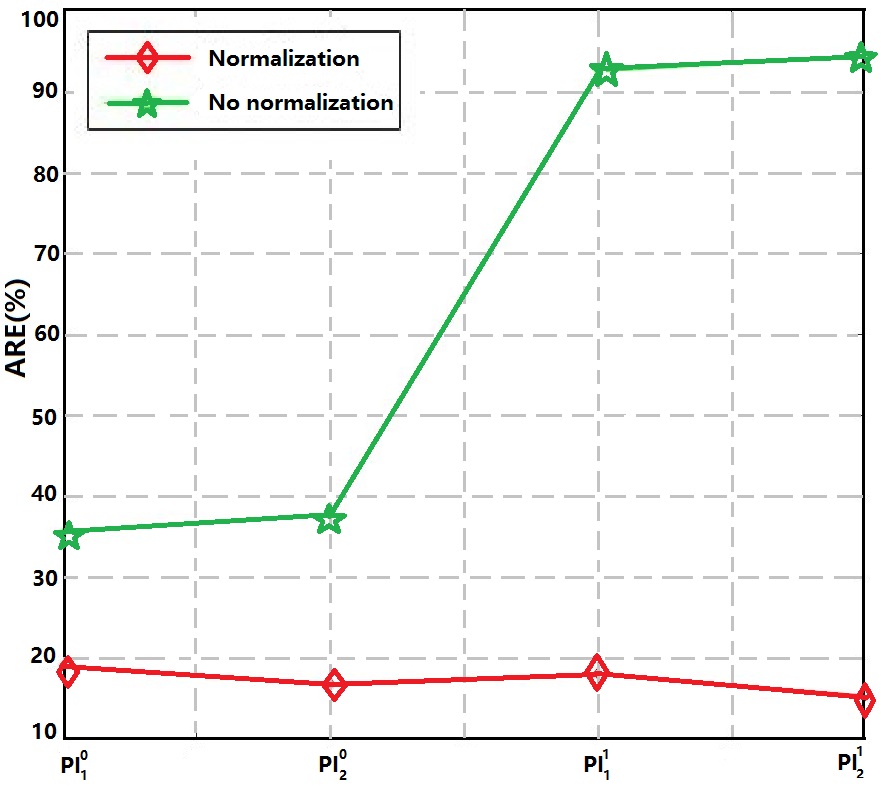}\\
  \caption{The calculation errors of $PIs$ , which are obtained by using the normalized method and not using the normalized method.}\label{Fig:6}
\end{figure}

In order to observe the property of $PIs$ more clearly, we use Chi-Square distance to calculate the feature distance between any two images. So, we can get a $25 \times 25$ distance matrix which is shown in Fig~\ref{Fig:5}. As the distance increases, the color changes from white to black. Obviously, the color of the area near the diagonal is lighter than that of other regions, indicating that $PIs$ of similar images are similar in value, and vice versa. When the number of pixels is not normalized, we also calculate the numerical values of $PIs$ of 25 test images by setting the neighborhood size to $9 \times 9$ $(\sigma=1.33)$. As shown in Fig~\ref{Fig:6}, by comparing  $ARE$ of $PIs$, we can find that the normalized method greatly reduces the computational error.

\subsection{Experiments on Synthetic Image Database}\label{sec:6.3}
\begin{figure}
  \centering
  \subfloat[20 kinds of butterfly images.]{\label{Fig:7a}\includegraphics[height=4.2cm,width=7cm]{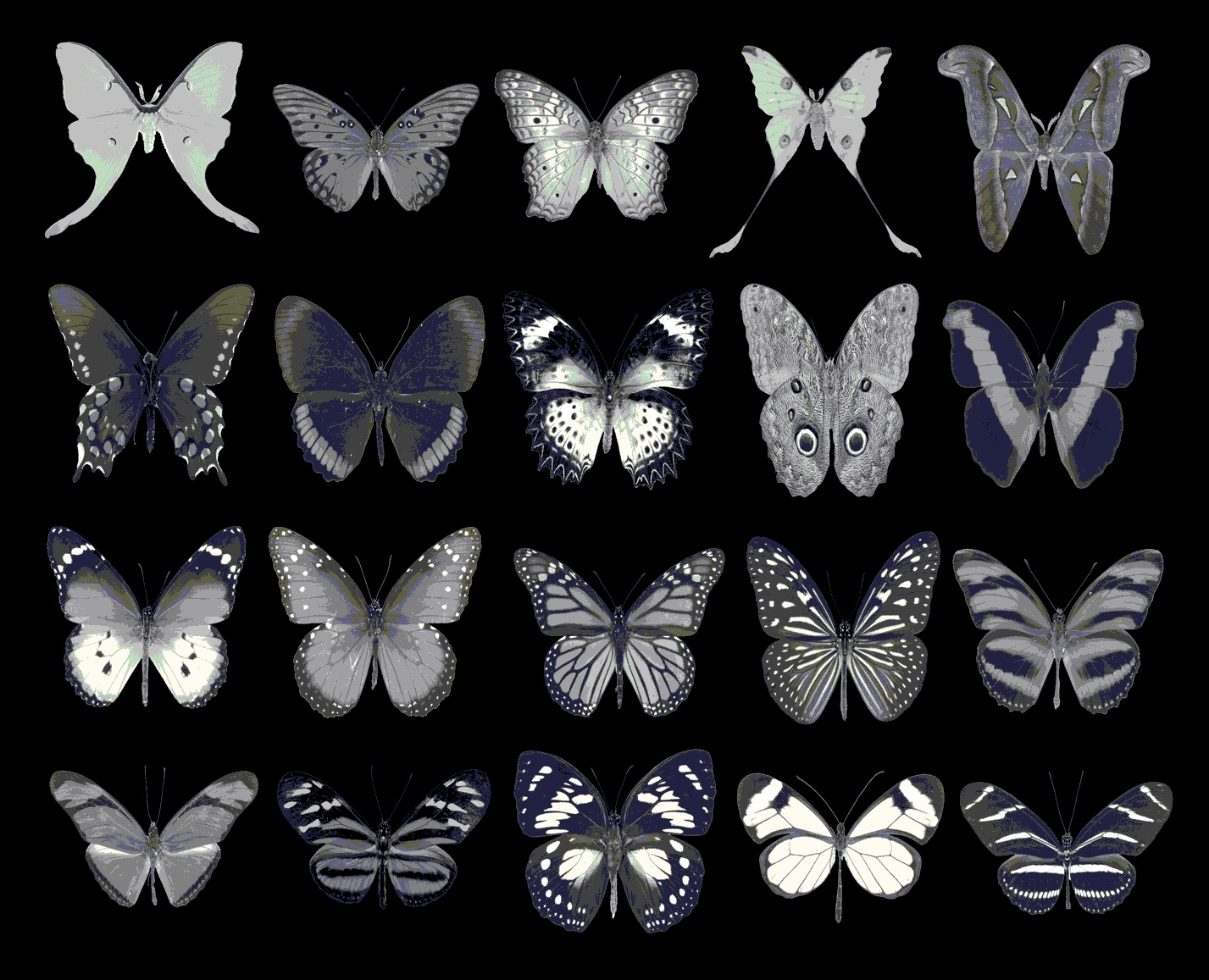}}\\
  \subfloat[Each image is transformed by 10 general projective transformations.]{\label{Fig:7b}\includegraphics[height=4.2cm,width=7cm]{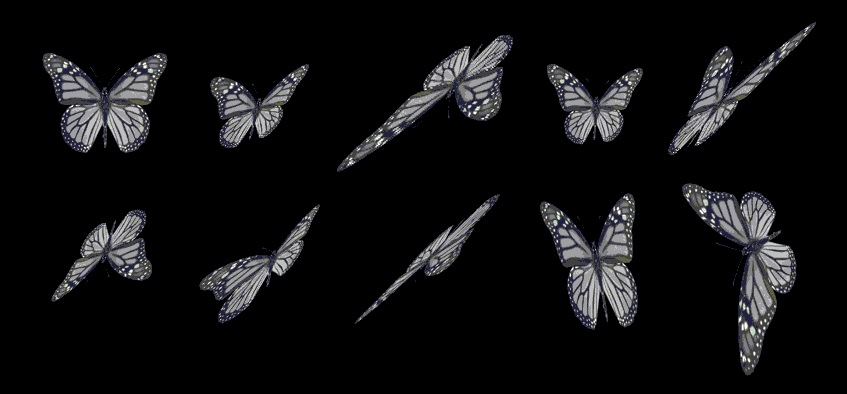}\hfill}
  \caption{Sample images from the synthetic image database}\label{Fig:7}
\end{figure}

Subsequently, we conduct retrieval experiments on a synthetic image database. 20 kinds of butterfly images are collected from the Internet, which are shown in Fig~\ref{Fig:7a}. Each image is transformed by 10 general projective transformations, as shown in Fig~\ref{Fig:7b}. So, there are 200 images in the synthetic image database.

For comparison, we choose 4 traditional moment invariants, affine moment invariants $(AMIs)$, Hu moments $(HMs)$, Zernike moments $(ZMs)$ and Gaussian-Hermite moments $(GHMs)$.
\begin{enumerate}
  \item $AMIs$: We choose $(AMI_{1}, AMI_{2},AMI_{3},AMI_{6},\\AMI_{7},AMI_{8},AMI_{9})$ proposed in \cite{18}, which are invariant to the affine transformation.
  \item $HMs$: $(HM_{1}, HM_{2}, HM_{3}, HM_{4}, HM_{5}, HM_{6}, HM_{7})$ proposed in \cite{5}, which are invariant to the similarity transformation.
  \item $ZMs$: $(Z_{11},Z_{2,0},Z_{2,2},Z_{3,1},Z_{3,3},Z_{4,0},Z_{4,2})$ proposed in \cite{6}, which are invariant to the similarity transformation.
  \item $GHMs$: $(\psi_{1}, \psi_{2}, \psi_{3}, \psi_{4}, \psi_{5}, \psi_{6}, \psi_{7})$ proposed in \cite{31}, which are invariant to rotation and translation.
\end{enumerate}

\begin{figure}
  \centering
  \includegraphics[height=7.5cm,width=8.3cm]{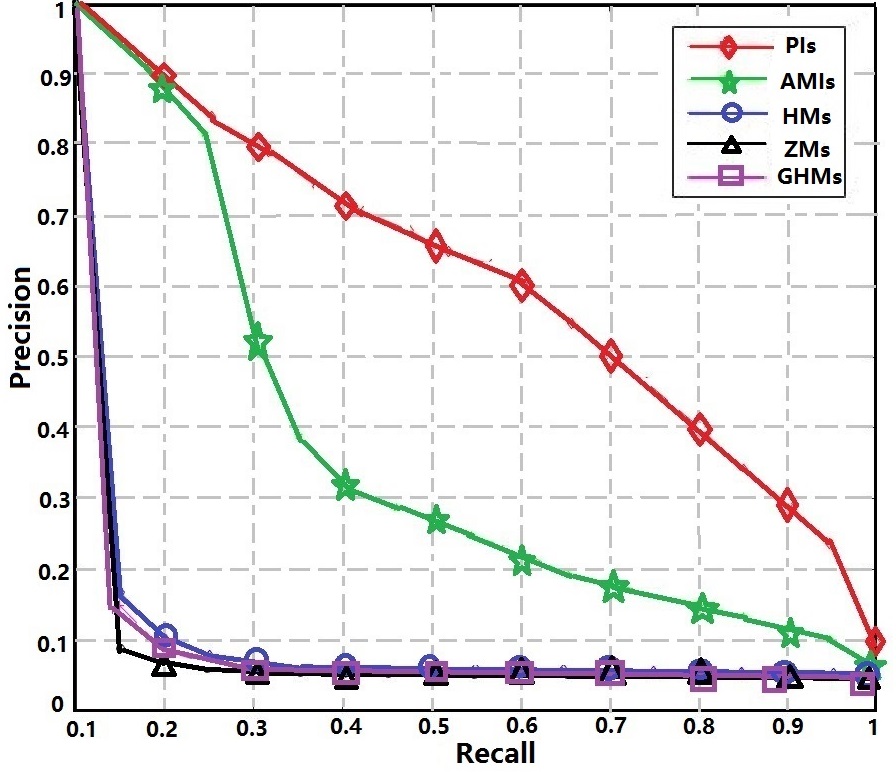}\\
  \caption{The Precision-Recall curves of $PIs$, $AMIs$, $HMs$, $ZMs$ and $GHMs$ on the synthetic image database}\label{Fig:8}
\end{figure}

 We make $PIs=(AMI_{1}, AMI_{2},AMI_{3}, PI^{0}_{1}, PI^{0}_{2}, PI^{1}_{1}, \\PI^{1}_{2})$ to ensure the consistency of the feature dimension. Also, the Chi-Square distance is used to calculate the feature distance between two images. We retrieval each image and draw 5 Precision-Recall curves in Fig~\ref{Fig:8}, which are obtained by using $PIs$, $AMIs$, $HMs$, $ZMs$ and $GHMs$.

 The Precision and Recall are defined by
 \begin{equation}\label{equ:59}
   Precision=\frac{|\{relavant~images\}\cap\{retrieved~images\}|}{\{retrieved~images\}}
 \end{equation}
 \begin{equation}\label{equ:60}
   Recall=\frac{|\{relavant~images\}\cap\{retrieved~images\}|}{\{relavant~images\}}
 \end{equation}

 As shown in Fig~\ref{Fig:8}, the retrieval result obtained by using $PIs$ is better than those obtained by other traditional moment invariants. This shows that the new features have good properties for general projective transformations. When the distance between the camera and the object is much larger than the size of the object itself, the projective transformation of the object can be represented by the affine transformation. But in this experiment, this condition is no longer satisfied. So, the retrieval result obtained by using $AMIs$ is poor. $HMs$, $ZMs$ and $GHMs$ are invariant to rotation, translation and scaling. When the object has a serious geometric deformation which can be represented by the general projective transformation, these moment invariants almost fail.

\subsection{Leaf Classification}
In order to further verify the performance of $PIs$, we choose some real image databases for testing. The image in leaf databases is generally removed from the background and has the single target, so it is well suitable for shape analysis. In this paper, we choose the Flavia database  proposed by Wu et~al. in \cite{27}. This database can be downloaded from http://flavia.sourceforge.net/ and contains 32 different kinds of leaves. Each category has about 60 images, a total of 1907 images.

\begin{figure*}
  \centering
  \subfloat[20 species leaves from the Flavia database]{\label{Fig:9a}\includegraphics[height=4.3cm,width=7.5cm]{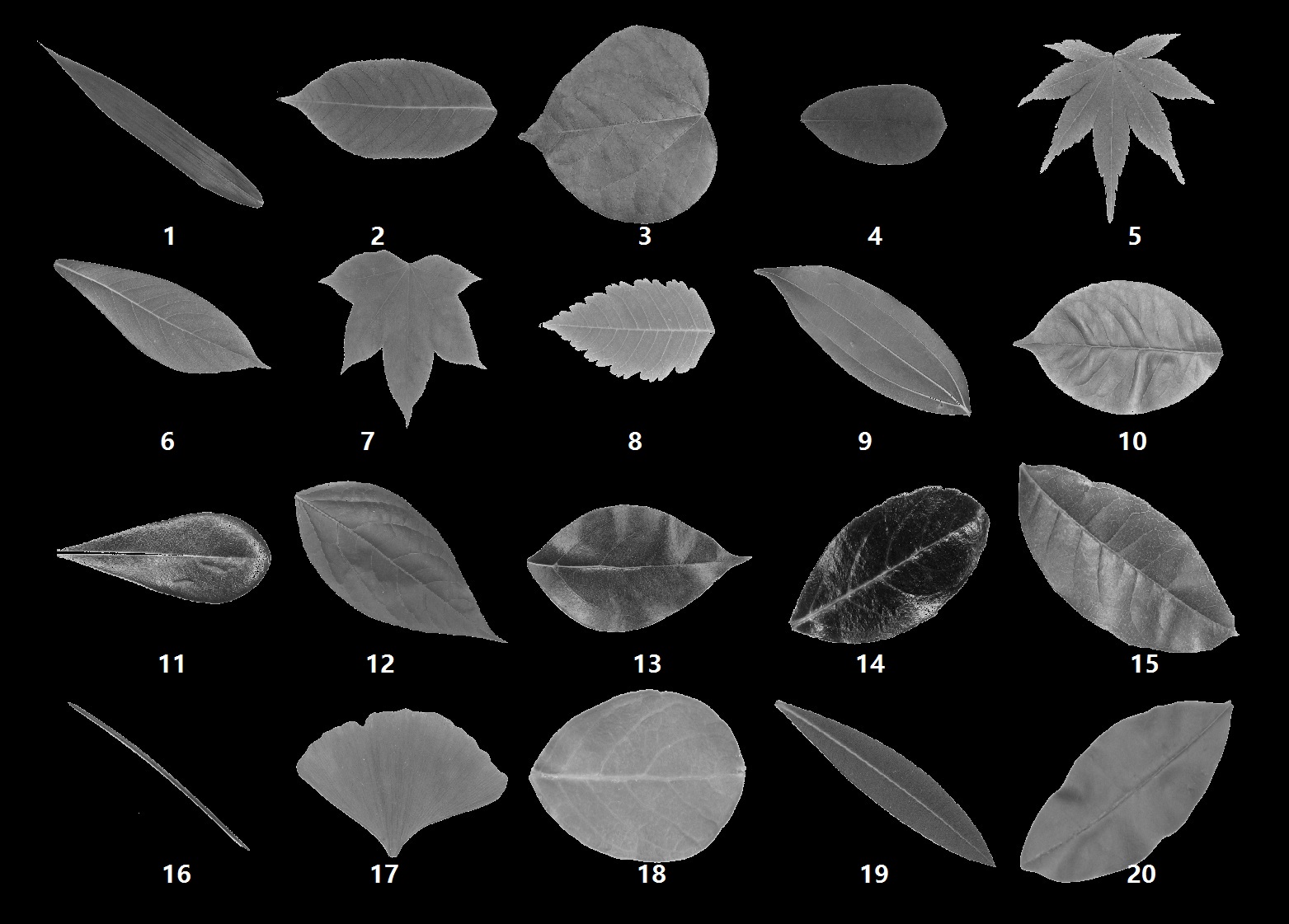}\hfill}~~~~~~~~~~~~
  \subfloat[Each classification has 60 images]{\label{Fig:9b}\includegraphics[height=4.3cm,width=7.5cm]{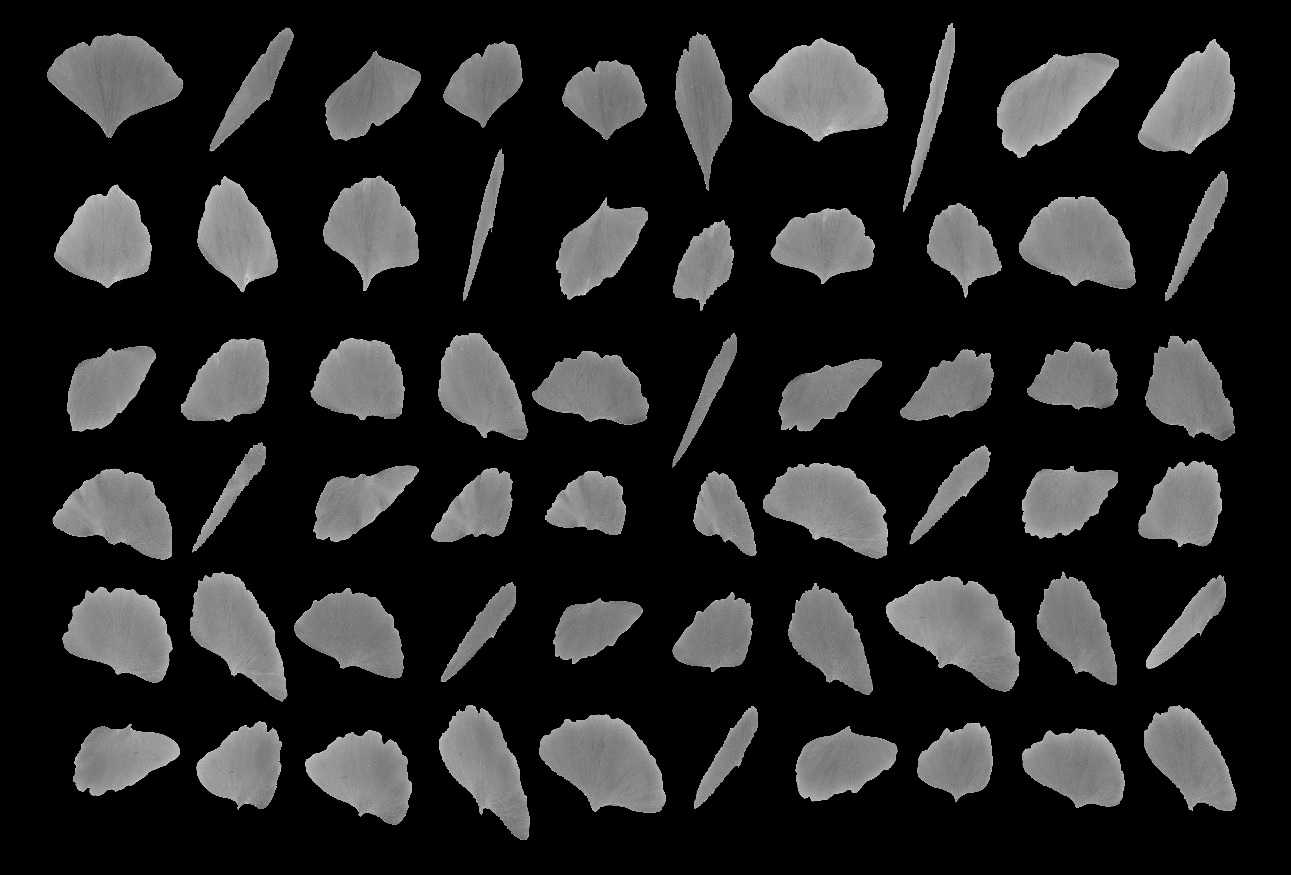}}
  \caption{Sample images from leaves database}\label{Fig:9}
\end{figure*}

Firstly, we select 20 species leaves from the Flavia database, which are shown in Fig~\ref{Fig:9a}. Then, we choose 10 images from each class. Each image is transformed by 6 general projective transformations. Thus, each species has 60 images, as shown in Fig~\ref{Fig:9b}. Finally, in order to carry out image classification experiment, $10\%$ images are selected randomly to be the training data and the rest 1080 images make up the testing data.

\begin{figure}
  \centering
  \includegraphics[height=8cm,width=8cm]{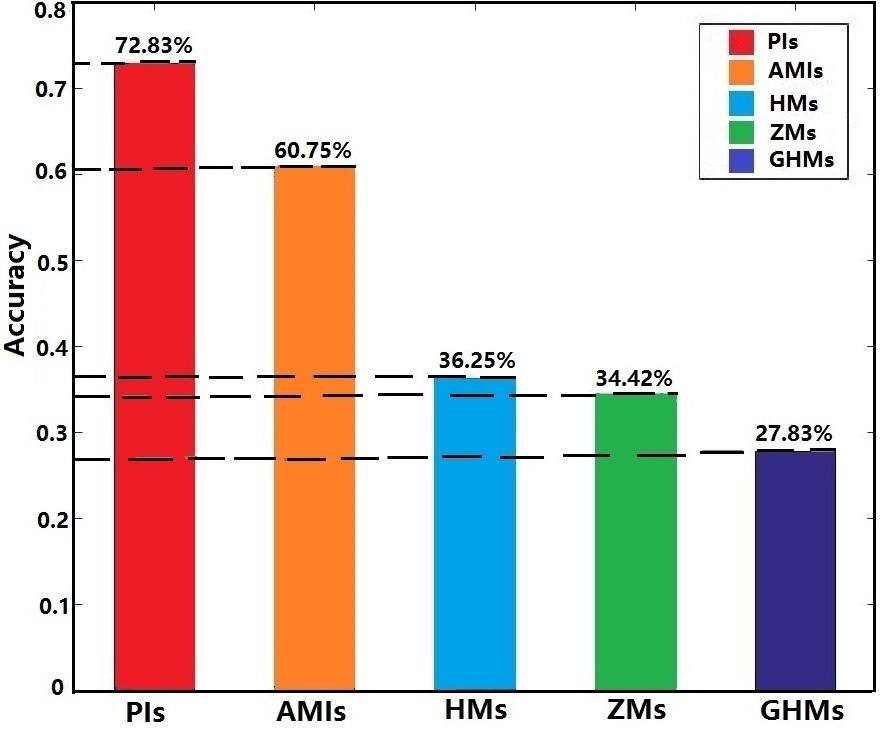}\\
  \caption{The classification accuracy of the whole leaves database}\label{Fig:10}
\end{figure}

We use the Nearest Neighbor classifier based on the Chi-Square distance to estimate the categories of the test images. The classification accuracy of the whole leaf database is shown in Fig~\ref{Fig:10}. And the classification accuracy of each species of leaves is listed in Table.~\ref{Tab:3}. The result show that the classification accuracy $(72.83\%)$ obtained by using $PIs$ is higher than those obtained using other moment invariants $(AMIs:60.75\%,~HMs:36.25\%,~ZMs:34.42\%,~GHMs:27.83\%)$. For 20 kinds of leaves, the highest classification accuracy of 15 categories is obtained by using $PIs$.

\begin{table}
\caption{The classification accuracy of each species of leaves.}
\centering
\begin{tabular}{p{5mm}<{\centering}p{11.3mm}<{\centering}p{11.3mm}<{\centering}p{11.3mm}<{\centering}p{11.3mm}<{\centering}p{11.3mm}<{\centering}}
  \toprule[1.1pt]
  Species & PIs & AMIs & HMs & ZMs & GHMs \\
  \toprule[1.1pt]
  1 & \textbf{85.0\%} & 61.7\% & 83.3\% & 51.7\% & 48.3\% \\
  \hline
  2 & \textbf{50.0\%} & 46.6\% & 31.7\% & 20.0\% & 21.7\% \\
  \hline
  3 & \textbf{65.0\%} & 51.7\% & 26.7\% & 18.3\% & 23.3\% \\
  \hline
  4 & 93.3\% &\textbf{100.0\%} & 33.3\% & 83.3\% & 18.3\% \\
  \hline
  5 & \textbf{93.3\%} &58.3\% & 20.0\% & 48.3\% & 25.0\% \\
  \hline
  6 & 36.7\% &\textbf{55.0\%} & 38.3\% & 30.0\% & 30.0\% \\
  \hline
  7 & \textbf{55.0\%} & 43.3\% & 23.3\% & 28.3\% & 23.3\% \\
  \hline
  8 & \textbf{91.7\%} & 78.3\% & 31.7\% & 43.3\% & 23.3\% \\
  \hline
  9 & \textbf{78.3\%} & 43.3\% & 25.0\% & 16.7\% & 20.0\% \\
  \hline
  10 & \textbf{65.0\%} & 50.0\% & 30.0\% & 20.0\% & 26.7\% \\
  \hline
  11 & \textbf{85.0\%} & 75.0\% & 31.7\% & 46.7\% & 25.0\% \\
  \hline
  12 & \textbf{58.3\%} & 35.0\% & 21.7\% & 21.7\% & 25.0\% \\
  \hline
  13 & \textbf{68.3\%} & \textbf{68.3\%} & 26.7\% & 31.7\% & 18.3\% \\
  \hline
  14 & \textbf{96.7\%} & 70.0\% & 35.0\% & 33.3\% & 26.7\% \\
  \hline
  15 & \textbf{81.7\%} & 41.7\% & 23.3\% & 18.3\% & 25.0\% \\
  \hline
  16 & 93.3\% & 81.6\% & \textbf{100\%} & 18.3\% & 25.0\% \\
  \hline
  17 & 65.0\% & \textbf{71.7\%} & 30.0\% & 20.0\% & 20.0\% \\
  \hline
  18 & \textbf{96.7\%} & 90.0\% & 33.3\% & 21.7\% & 33.3\% \\
  \hline
  19 & \textbf{60.0\%} & 36.7\% & 51.7\% & 30.0\% & 25.0\% \\
  \hline
  20 & 40.0\% & \textbf{58.3\%} & 28.3\% & 18.3\% & 21.7\% \\
  \toprule[1.1pt]
\end{tabular}
\label{Tab:3}
\end{table}

\subsection{Object Retrieval}
\begin{figure*}
  \centering
  \subfloat[20 different classes of objects in COIL-20.]{\label{Fig:11a}\includegraphics[height=4.2cm,width=7.5cm]{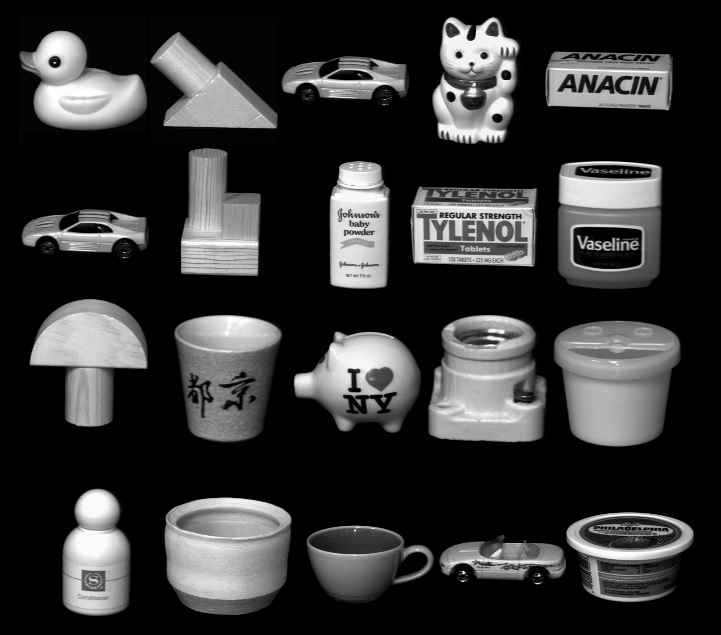}\hfill}~~~~~~~~~~~~
  \subfloat[Each object is photographed from 72 different angles.]{\label{Fig:11b}\includegraphics[height=4.2cm,width=7.5cm]{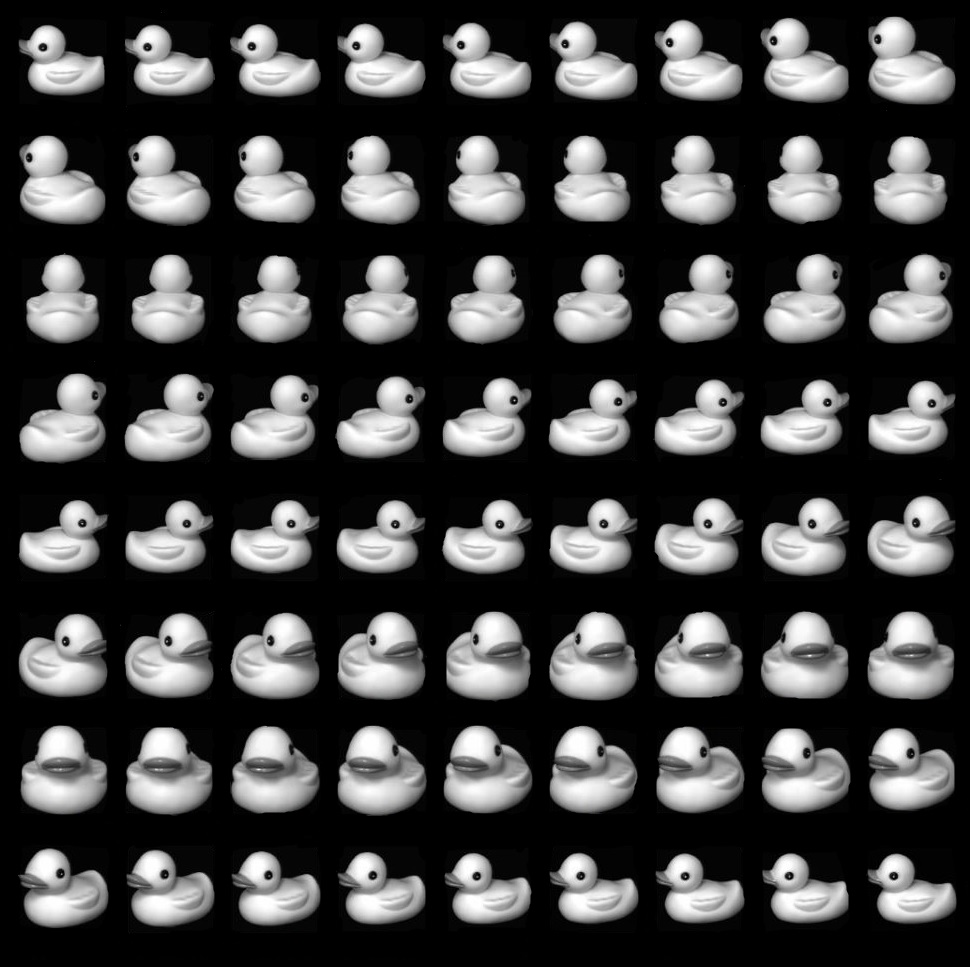}\hfill}
  \caption{Sample images from COIL-20.}\label{Fig:11}
\end{figure*}

Another database is COIL-20 in \cite{12}. It contains 1440 images belonging to 20 different classes of objects which are shown in Fig~\ref{Fig:11a}. As shown in Fig~\ref{Fig:11b}, each object is photographed from 72 different angles.

\begin{figure}
  \centering
  \includegraphics[height=8cm,width=8.3cm]{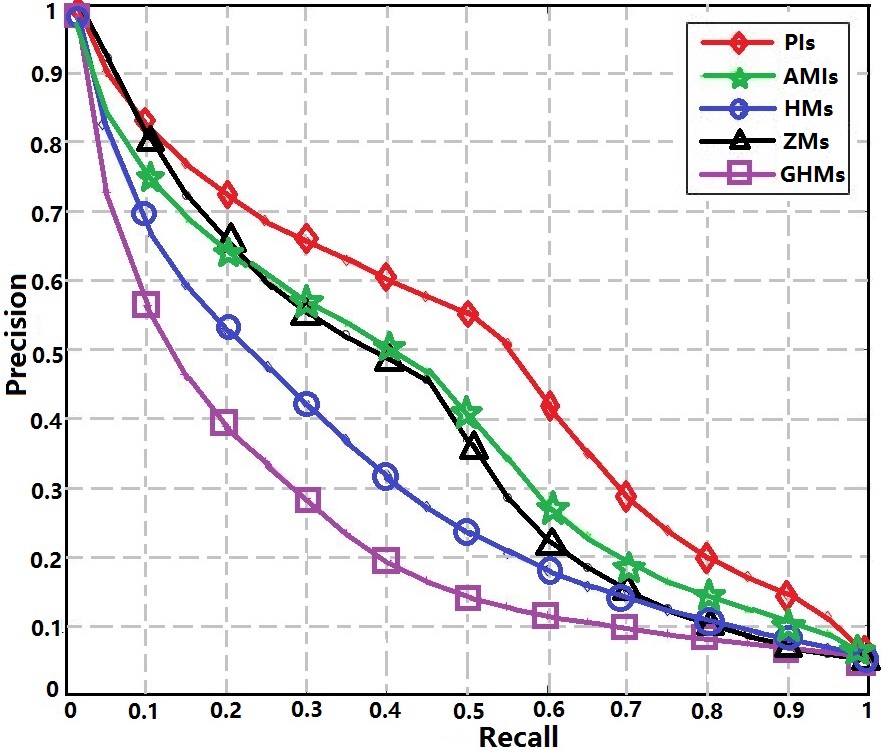}\\
  \caption{The Precision-Recall curves of $PIs$, $AMIs$, $HMs$, $ZMs$ and $GHMs$ on COIL-20}\label{Fig:12}
\end{figure}

Similar to Section \ref{sec:6.3}, the image retrieval experiment is performed on this database. 5 Precision-Recall curves are plotted in Fig~\ref{Fig:12}, which are obtained by using $PIs$, $AMIs$, $HMs$, $ZMs$ and $GHMs$. It can be found that for the real image database, the retrieval result obtained by using $PIs$ is better than those obtained by other moment invariants, too.

This is because the real geometry deformation caused by the change of the viewpoint is close to the general projective transformation. So, the features that have invariance for projective transformations are more advantageous than others.

\begin{figure*}
  \centering
  \subfloat[$PIs$]{\label{Fig:13a}\includegraphics[height=4.2cm,width=7.7cm]{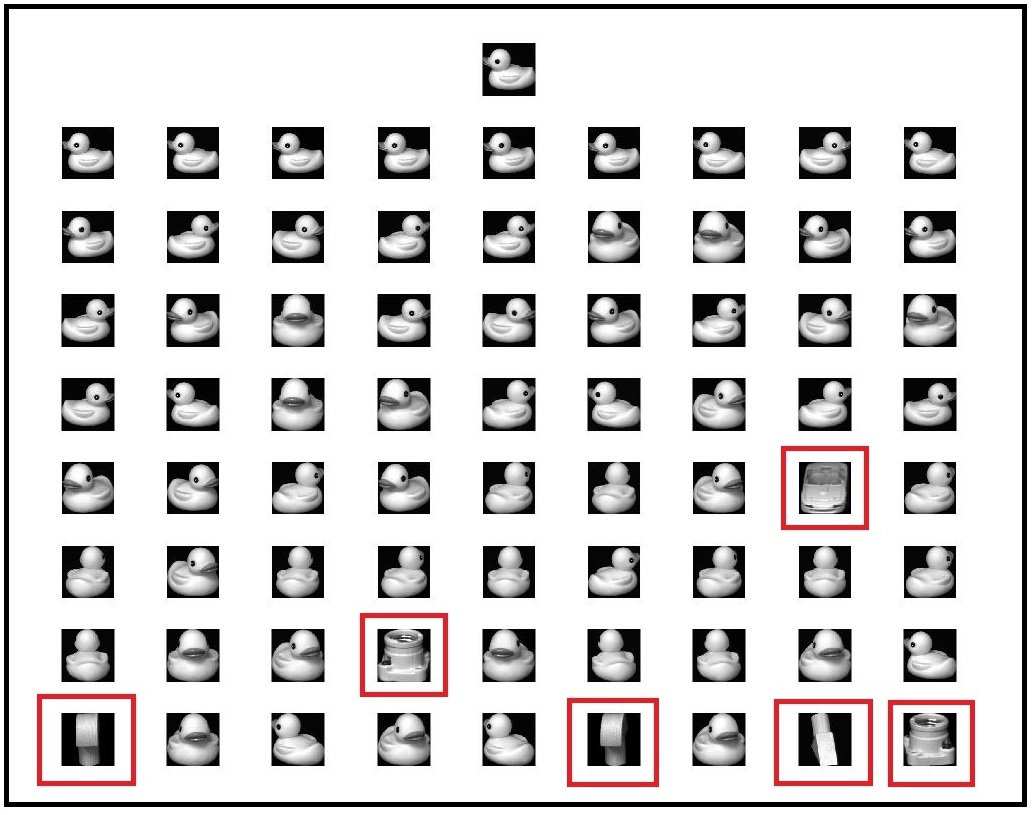}\hfill}~~~~~~~~~~~
  \subfloat[$AMIs$]{\label{Fig:13b}\includegraphics[height=4.2cm,width=7.7cm]{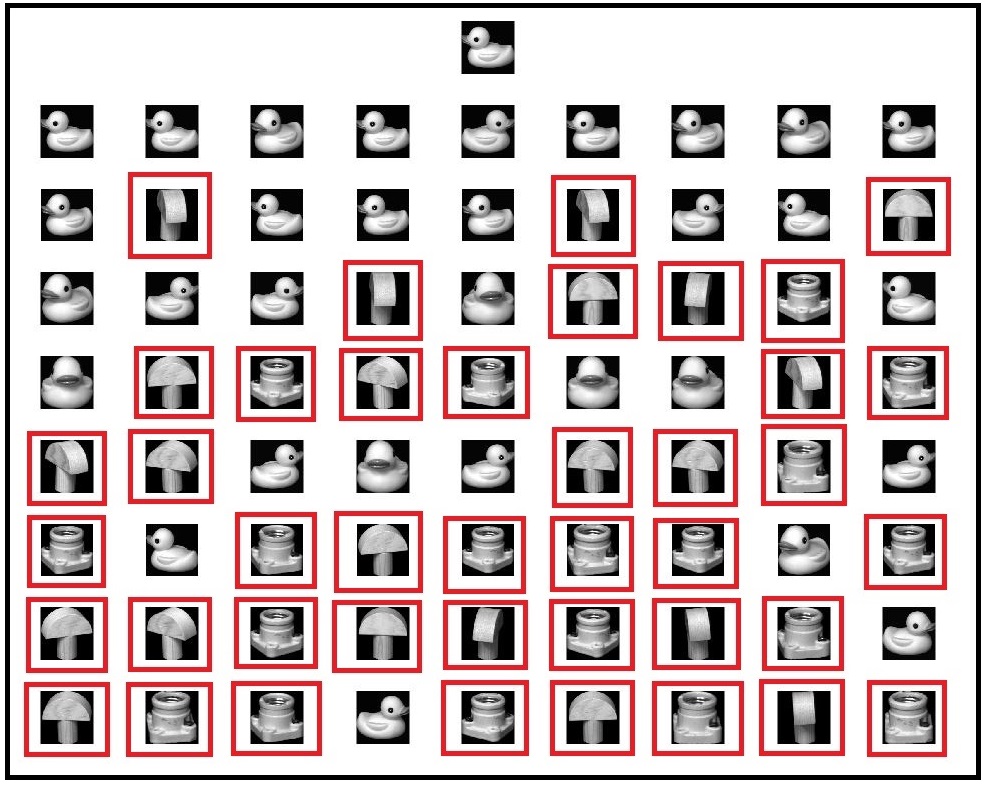}}
  \caption{Two retrieval results for the same image, which are obtained by using $PIs$ and $AMIs$. The red box indicates the wrong retrieval results.
}\label{Fig:13}
\end{figure*}

In order to explain the retrieval result more clearly, we give two retrieval results for the same image, which are obtained by using $PIs$ and $AMIs$. We calculate the feature distance between every image in COIL-20 and the test image, and arrange all images in ascending order according to their feature distances, and give the first 72 images in Fig~\ref{Fig:13}. As shown in Fig~\ref{Fig:13a}, only 6 images belonging to other categories are retrieved by using $PIs$. However, 41 images belonging to other categories are retrieved by using $AMIs$, which are shown in Fig~\ref{Fig:13b}. It has been proved that, as a kind of moment invariants, $PIs$ perform better than other traditional moment invariants and are robust for the real geometric deformation.

\section{Conclusion}\label{sec:7}
This paper presents general ways of building $RPDIs$ and $PIs$, which provide two promising tools in shape analysis, and could be expected a better performance in describing images and broader applications. They break the common belief that no projective invariants exist. By using $PIs$, they avoid the problem of solving the transcendental function. Instead, they eliminates the non-constant Jacobian with corresponding $RPDIs$. Meanwhile, it's unnecessary to set additional constraints to projective parameters. Also, it could be applied to images without knowing corresponding points in advance. With $PIs$, one can compare the similarity between images under the projective transformation without knowing the parameters of the transformation, which provides a good tool to shape analysis in image processing, computer vision and pattern recognition.

Then, we find that partial derivatives of the discrete image and the change in the number of pixels before and after the geometric transformation are the main reasons causing the calculation error of $PIs$. After comparing various methods, we employ derivatives of the Gaussian function as filters to compute partial derivatives of the discrete image via convolution and design the normalization method to reduce the error caused by the change in the number of pixels. By using these methods, the properties of PMIs are greatly improved. So, it's possible to use these invariants in practical applications.

Finally, some experiments are designed to evaluate the invariance and robustness of PIs. The results of image classification and retrieval show that PIs have better performance than other moment invariants on the image databases, where the same kind of images satisfy various projective transformation relations.
%\appendices

%\section{}

\ifCLASSOPTIONcompsoc
  \section*{Acknowledgments}
\else
  \section*{Acknowledgment}
\fi
This work has been funded by National Natural Science Foundation of China (Grant No. 60873164, 61227802 and 61379082).
\ifCLASSOPTIONcaptionsoff
 \newpage
\fi

\end{document}